\documentclass[10pt,letterpaper]{article}
\usepackage[top=0.85in,left=2.75in,footskip=0.75in]{geometry}

\usepackage{amsmath,amssymb}

\usepackage{changepage}
\usepackage{booktabs}
\usepackage{hhline}

\usepackage{textcomp,marvosym}

\usepackage{cite}

\usepackage{nameref,hyperref}

\usepackage[right]{lineno}
\newlength{\Oldarrayrulewidth}
\newcommand{\Cline}[2]{%
  \noalign{\global\setlength{\Oldarrayrulewidth}{\arrayrulewidth}}%
  \noalign{\global\setlength{\arrayrulewidth}{#1}}\cline{#2}%
  \noalign{\global\setlength{\arrayrulewidth}{\Oldarrayrulewidth}}}
\usepackage[nopatch=eqnum]{microtype}
\DisableLigatures[f]{encoding = *, family = * }

\usepackage[table]{xcolor}

\usepackage{array, boldline, makecell, booktabs}
\usepackage{soul}
\usepackage[normalem]{ulem}

\newcolumntype{+}{!{\vrule width 2pt}}

\newlength\savedwidth

\newcommand{\fmicro}{F1\textsubscript{micro} }
\newcommand{\fmacro}{F1\textsubscript{macro} }
\newcommand{\Fmicro}{F1\textsubscript{micro} }
\newcommand{\Fmacro}{F1\textsubscript{macro} }

\usepackage{setspace} 
\raggedright
\usepackage[aboveskip=1pt,labelfont=bf,labelsep=period,justification=raggedright,singlelinecheck=off]{caption}

\makeatletter
\renewcommand{\@biblabel}[1]{\quad#1.}
\makeatother

\usepackage{lastpage,fancyhdr,graphicx}
\usepackage{epstopdf}
\usepackage{tabularray}
\fancyheadoffset[L]{2.25in}
\fancyfootoffset[L]{2.25in}
\usepackage{multirow}

\usepackage{float}

\begin{document}
\vspace*{0.2in}

\begin{flushleft}
{\Large
\textbf\newline{Comparison between parameter-efficient techniques and full fine-tuning: A case study on multilingual news article classification} 
}
\newline
\\
Olesya Razuvayevskaya\textsuperscript{1\Yinyang*},
Ben Wu\textsuperscript{1\Yinyang},
João A. Leite\textsuperscript{1\Yinyang},
Freddy Heppell\textsuperscript{1\Yinyang},
Ivan Srba\textsuperscript{2},
Carolina Scarton\textsuperscript{1},
Kalina Bontcheva\textsuperscript{1},
Xingyi Song\textsuperscript{1}
\\
\bigskip
\textbf{1} Department of Computer Science, The University of Sheffield, Sheffield, UK \\
\textbf{2} Kempelen Institute of Intelligent Technologies, Bratislava, Slovakia
\\
\bigskip

%
%
\Yinyang These authors contributed equally to this work.

* o.razuvayevskaya@sheffield.ac.uk

\end{flushleft}
\section*{Abstract}
Adapters and Low-Rank Adaptation (LoRA) are parameter-efficient fine-tuning techniques designed to make the training of language models more efficient. Previous results demonstrated that these methods can even improve performance on some classification tasks. This paper complements existing research by investigating how these techniques influence  classification performance and computation costs compared to full fine-tuning. We focus specifically on multilingual text classification tasks (genre, framing, and persuasion techniques detection; with different input lengths, number of predicted classes and classification difficulty), some of which have limited training data. In addition, we conduct in-depth analyses of their efficacy across different training scenarios (training on the original multilingual data; on the translations into English; and on a subset of English-only data) and different languages. Our findings provide valuable insights into the applicability of parameter-efficient fine-tuning techniques, particularly for multilabel classification and non-parallel multilingual tasks which are aimed at analysing input texts of varying length.


\section*{Introduction}
The development of language models has led to a significant increase in the number of trainable parameters needed to fine-tune such models, with state-of-the-art models comprising of millions or even billions of parameters \cite{devlin2018bert,raffel2020exploring}. This poses a serious constraint on the process of fine-tuning such models, often relying on significant computational resources. Many of the recent research efforts are therefore focused on the development of more efficient training techniques \cite{zaken2021bitfit, jiang2019smart, xu2021raise}.
Methods that can decrease the computational costs make language models more accessible to researchers and practitioners with limited computational resources, and reduces the carbon footprint of their training.

In this study, we investigate the effectiveness of parameter-efficient fine-tuning techniques (PEFTs) in multilingual, monolingual and cross-lingual text classification scenarios. We have included PEFT techniques that have been evaluated on large volumes of data ($>$20B) in prior research \cite{lialin2023scaling} -- namely, LoRA, adapter, BitFit, prefix tuning and prompt tuning. Since the majority of these techniques were previously evaluated on transformer-based models (BERT and RoBERTa) for text classification tasks, we select the subset of those that can be applied to such models -- namely LoRA, adapter and BitFit.
We exclude prefix tuning \cite{li2021prefix} and prompt tuning \cite{lester2021power} as these techniques are  applied to generative Large Language Models (LLMs) models for text-to-text generation tasks and cannot be directly compared to the above-mentioned three methods.

In more detail, adapter-based fine-tuning represents a family of efficiency techniques that work by freezing a pre-trained language model and adding a small number of trainable parameters 
in the layers of the language model \cite{bapna-firat-2019-simple, houlsby2019parameter, he-etal-2021-effectiveness}. This significantly reduces the training time at the cost of a small or no performance penalty. Another method of reducing the number of trainable parameters is based on performing Low-Rank Adaptation (LoRA) \cite{hu2021lora}. The main idea behind the LoRA approach is to freeze the weights of pre-trained language models and insert low-rank decomposition matrices into the transformer layers. BIas-Term FIne-Tuning (BitFit) is a PEFT technique that uses only the bias term and the task-specific linear classification layer during training, omitting most of the parameters in the encoder-decode layers \cite{zaken2021bitfit}. This results in a significant reduction in computational costs as bias terms represent only a small fraction (up to 0.1\%) of all the parameters of the models.

Prior studies report that, in addition to reducing computational costs, adapter-based methods can outperform full fine-tuning (FFT) in zero-shot cross-lingual settings\cite{he-etal-2021-effectiveness, chalkidis-etal-2021-multieurlex, xenouleas2022realistic}. However, the tasks addressed in these studies focus on parallel multilingual texts \cite{chalkidis-etal-2021-multieurlex}, meaning that inference is not performed in a realistic zero-shot manner \cite{xenouleas2022realistic}, on short inputs \cite{he-etal-2021-effectiveness} or in highly specific domains \cite{xenouleas2022realistic}. Given that adapter methods are additionally known to have limited capabilities in processing long text inputs due to reserving a part of the sequence length for
adaptation \cite{hu2021lora}, there is a motivation for evaluating this method in non-parallel cross-lingual and multilingual settings and on different input lengths. LoRA, on the other hand, does not reduce the input sequence length, and prior comparisons between FFT and LoRA suggest that, in addition to being parameter-efficient, LoRA can, for certain models, outperform FFT \cite{hu2021lora}. However, there is currently a research gap in terms of evaluating LoRA in similar multilingual training scenarios, which creates a motivation for including this PEFT in our study. There is no prior evidence regarding the cross-lingual capabilities of BitFit, however, this PEFT technique was shown to have comparable performance to adapters and to also be comparable to FFT on the GLUE benchmark. This motivated the inclusion of the BitFit method in this research.

This study is therefore motivated by the lack of consistent evaluation of PEFT techniques in multilingual and cross-lingual zero-shot classification tasks performed on long non-parallel texts. This gap in current research is addressed through a systematic comparative investigation of how adapter, LoRA and FitBit techniques perform on  multilingual multilabel  classification tasks, both in terms of classification performance and computation costs. We first perform an ablation study on one classification task to identify the PEFT techniques that demonstrate the best performance in multilingual and cross-lingual scenarios. Those best performing PEFT methods are then investigated further in three training scenarios (multilingual and cross-lingual) on 
three diverse classification tasks.

In particular, we study the behaviour of these PEFTs on three multilingual multilabel news article classification tasks introduced as separate sub-tasks of the recent SemEval 2023 Shared Task 3 \cite{semeval2023task3}: news genre, framing and persuasion technique detection. All three tasks are based on multilingual data and contain `unseen' languages, i.e. languages not available in the training set, but present in the test set. Further motivation for this research came from the success of our three original best performing approaches in each of these three different sub-tasks\cite{wu2023sheffieldveraai, hromadka2023kinitveraai}. 
Our best method for sub-task 1 was based on an ensemble of FFT and adapter-based models,
as well as language-specific checkpoint selection. Our best sub-task 2 approach was based on mono- and multilingual ensembles, one of which combined FFT and adapter methods with task adaptive pre-training \cite{gururangan2020don}. Finally, our models for sub-task 3 included language-specific classification threshold selection and the incorporation of unlabelled data into the training corpus. 
Overall, we found that adapters improved performance in certain monolingual scenarios in sub-task 1 and for multilingual ones in sub-task 2.

The main contributions of this paper are the following:

\begin{itemize}
    \item \textbf{Provide an evaluation of PEFTs on diverse text classification tasks} as compared to fully fine-tuned models.
    \item \textbf{Compare the computational costs for training FFT and PEFT models on the three sub-tasks as such an evaluation has not been performed before on these datasets and tasks.} We show that PEFTs significantly reduce the number of trainable parameters (between 140 and 280 times less parameters) and achieve shorter training times (between 32\% and 44\%). Unlike previous analyses between bottleneck adaptors, LoRA and FFT, our comparison is novel is providing more fine-grained statistics, such as the peak VRAM usage and the relative train time duration. The latter statistic is particularly important because the training process 
typically more time-consuming than inference.
    \item \textbf{Carry out an in-depth comparison of the performance of PEFTs in different training scenarios}, investigating 
    both joint multilingual training and two types of single source language training scenarios. We evaluate how each method performs on seen languages and generalises to unseen ones.
    
    \item \textbf{Improve on the original highest SemEval 2023 results.} 
    For sub-task 3, we achieved a better performance on eight out of the nine languages compared to the top results in the official leaderboard. For sub-tasks 1 and 2 the results reported here are mostly comparable to our original SemEval 2023 submissions despite the fact that in this study we use significantly less complex models (the original solutions utilised multiple sub-task tailored steps and complex ensembles).

\end{itemize}

\section*{Related work} \label{sec:background}

\subsection*{Parameter-efficient fine-tuning techniques}

As already mentioned above, PEFTs are computationally efficient due to limiting the number of trainable parameters.
All such techniques freeze the pre-trained model, but differ in the location of the inserted trainable parameters. We focus specifically on Bottleneck adapters and on Low-Rank Adaptation (LoRA).

\textbf{Bottleneck adapters} \cite{houlsby2019parameter} have a structure similar to autoencoders. The transformer hidden state $h$ is first down-projected to some smaller dimensionality $d_\text{bottleneck}$ with matrix $W_{\mathrm{down}}$, passes through non-linearity function $f$, and is then up-projected to the original dimensionality with matrix $W_{\mathrm{up}}$, and a residual connection $r$. This is defined formally in Eq \ref{eq:bottleneck}.

\begin{equation}
    \label{eq:bottleneck}
    h \leftarrow W_{\mathrm{up}} \cdot f(W_{\mathrm{down}} \cdot h) + r
\end{equation}

The location of the adapter layer depends on the adapter type. Houlsby adapters \cite{houlsby2019parameter} place adapter layers after the multi-head attention and feed-forward block, whereas Pfeiffer adapters \cite{pfeiffer-etal-2020-mad} place the adapter layer only after the feed-forward block.
Although adding the extra layer reduces the number of trainable parameters and thus increases the speed of fine-tuning, it also increases the number of overall parameters permanently, slowing down inference.

\textbf{LoRA} \cite{hu2021lora} adds low-rank decomposition matrices to the query ($Q$), key($K$), value ($V$) and pre-trained $W_0$ matrices of the self-attention sub-layers of the transformer.

Given a layer expressed as the matrix multiplication $h \leftarrow W_0x$ (where $W_0 \in \mathbb{R}^{d \times k}$), during the fine-tuning process, the value of $W_0$ is modified by some $\Delta W$. LoRA represents this delta as the low-rank decomposition $\Delta W = BA$ (where $B \in \mathbb{R}^{d\times r}, A \in \mathbb{R}^{r \times k}$) of rank $r \ll \min (d,k)$. Here, $W_0$ is frozen, while $B$ and $A$ are initialised randomly and updated during fine-tuning. The decomposition is scaled by hyperparameter $\alpha$ and rank $r$, thus giving the new expression in Equation~\ref{eq:lora}.

\begin{equation}
    \label{eq:lora}
    h \leftarrow W_0x + \frac{\alpha}{r} B A x
\end{equation}

Once the fine-tuning stage has been completed, the additional matrices can be removed by simplifying $W_0$, $A$ and $B$ to a single matrix $W_0^\prime$, thus giving the same number of parameters as the original pre-trained model. This solves the problem of increased inference time. When performing hyperparameter search, the $\alpha$ hyperparameter can be fixed since it is proportional to the learning rate \cite{dettmers2023qlora}.

Transformer models consist of six weight matrices, $W_0$, $W_K$, $W_V$, $W_Q$, and two matrices in multilayer perceptron (MLP) layer. In principle, it is possible to adapt any number of weight matrices, however, the authors claim that adapting only self-attention matrices ($W_0$, $W_K$, $W_V$, $W_Q$) produces results comparable to adapting all layers \cite{hu2021lora}.

Evaluating using LoRA adaptation in other parts of the model (all attention layers, all feed-forward layers, all layers, and attention and feed-forward output layers) revealed that inserting LoRA adaptation in all layers results in the highest performance, and that in this configuration the hyperparameter $r$ has no effect \cite{dettmers2023qlora}.

\textbf{BitFit} \cite{zaken2021bitfit} uses only a small percentage of model parameters called \textit{bias terms}. This allows a significant reduction in the number of trainable parameters. 

Each self-attention head $m$ at layer $l$ consists of the key (K), query(Q) and value (V) linear layers. Unlike in the FFT fine-tuning method, only the bias terms $b^l$ are considered during fine-tuning:

\begin{equation}
\begin{aligned}
    &Q^{m,l}(x)=W^{m,l}_qx+b^{m,l}_q\\
     &K^{m,l}(x)=W^{m,l}_kx+b^{m,l}_k\\
      &V^{m,l}(x)=W^{m,l}_vx+b^{m,l}_v\\
      \end{aligned}
\end{equation}

Multiple attention heads are then combined using an attention mechanism and fed into the MLP with layer-norm (LN). The biases from the query, key and value layers, attention and normalisation layers are the only fine-tuned parameters of the network. The $W^{(l),\cdot}_\cdot$ matrices and other MLP parameters are kept frozen. This reduces the number of trainable parameters to 0.08\%--0.09\%. 

The authors additionally show how the number of trainable parameters can be further reduced by focusing only on the bias terms from the query layer and the second MLP layer.
\subsection*{Related work on the comparison of adapters, LoRA and BitFit to FFT}
Despite significantly reducing the number of trainable parameters, bottleneck adapters were previously found to have minimal negative impact on the performance of fine-tuned models for simple sentence classification tasks. In particular, the evaluation on the GLUE benchmark dataset \cite{wang-etal-2018-glue} (a collection of sentence and sentence-pair classification tasks) showed the bottleneck adapter performance to be within 0.8\% of the performance of FFT, whilst only training 3.6\% of parameters \cite{houlsby2019parameter}. Additionally, in the field of machine translation, Bapna and Firat \cite{bapna-firat-2019-simple} found that adapters produce equivalent or even better results compared to FFT.

Adapters received special attention in the context of  multilingual tasks, with various studies reporting a consistent advantage of adapter models in comparison to FFT. In particular, Pfeiffer et al. \cite{pfeiffer-etal-2020-mad} proposed a modular adapter-based architecture, which combines task-specific adapters in the source language with `language' adapters trained on unlabelled data in the target language using the masked language modelling (MLM) objective. The authors report that this framework is able to outperform the traditional fully fine-tuned language models in cross-lingual transfer inference for the majority of language pairs. Different from Pfeiffer et al. \cite{pfeiffer-etal-2020-mad}, He et al. \cite{he-etal-2021-effectiveness} investigated the zero-shot cross-lingual capabilities of adapters without additionally training them on unlabelled target language data. The authors found that adapters still outperform FFT on named entity recognition (NER), part of speech tagging (POS) tagging and cross-lingual Natural Language Inference (XNLI) tasks. They report that the former method is particularly beneficial in low-resource and cross-lingual tasks, since it mitigates forgetting effects by minimising the differences between the representations of the fine-tuned and the pretrained model.

Unlike previous studies that mainly focused on the analysis of short texts, Chalkidis et al. \cite{chalkidis-etal-2021-multieurlex} investigated the performance of adapter models on long legal documents. Consistently with previous works, the authors found that bottleneck adapters outperform FFT, and provide better zero-shot cross-lingual capability. Their findings are based on the MultiEURLEX dataset, which consists of 65,000 EU law texts in 23 languages, categorised at multiple levels of detail (between 21 and 567 categories). In some aspects, this dataset is comparable to the dataset for sub-task 3 that we explore in this work, as both datasets are multilabel and multilingual, and have a comparable number of labels at MultiEURLEX's lowest level. However the style of EU law is naturally much more rigid and very different to news articles. Additionally, due to the specific data collection methodology used for MultiEURLEX, this dataset is not likely to contain irrelevant text. Finally, the MultiEURLEX dataset contains parallel multilingual data, meaning that the model making a cross-lingual zero-shot prediction on the target language has already `seen' this text at training time in another language.

The latter limitation of the MultiEURLEX dataset became the focus of the study by Xenouleas et al. \cite{xenouleas2022realistic} who questioned whether the findings of Chalkidis et al. would generalise to non-parallel datasets. When the dataset is modified to include only non-parallel documents, the authors found that translation-based methods (translate-test and translate-train) outperform multilingual models. Consistently with previous research, however, the authors observed that adapters still outperform the FFT in each of the settings, cross-lingual and translation-based. We note that this is likely dependent on domain and on whether the relevant properties are significantly affected by translation: legal documents are more likely to be properly represented in the target language than, for example, the language-specific linguistic properties signalling certain persuasion techniques. Additionally, the joint multilingual experiments conducted by Xenouleas et al. \cite{xenouleas2022realistic} do not leave out any languages to perform the zero-shot cross-lingual inference, which makes this approach not comparable to the monolingual ones due to the difference in the size of the training set. 

Taking into account the limitations of prior research in terms of the input length, high specificity of the domains, and parallelism of multilingual data, we aim to perform a wider comparison of adapter models with FFT on a variety of tasks and on non-parallel input texts that are not limited to a specific domain and vary in length. Another important limitation of prior work is that all the studies mentioned above evaluate cross-lingual transfer capabilities in a one-to-many manner, while the joint multilingual training setting does not perform zero-shot cross-lingual inference, making it not comparable with the monolingual zero-shot cross-lingual scenarios. To fill this gap and bring understanding into how the joint multilingual training data can affect cross-lingual capabilities of the methods, we aim to introduce a many-to-many joint inference into our comparison scenarios while keeping certain languages as 'unseen' for all the training scenarios.

To our knowledge, no comparison of LoRA and BitFit to the FFT method in a similar multilingual scenario exists. However, given the prior evidence that LoRA and BitFit can, for certain tasks, outperform FFT \cite{hu2021lora, zaken2021bitfit}, there exists a clear motivation of performing a pioneering evaluation of these PEFT techniques in multilingual scenarios.
LoRA technique can be particularly promising for our objective due to its ability to perform better on longer texts than adapter methods due to the fact that LoRA does not reduce the input sequence length \cite{hu2021lora}.

\section*{Dataset and Task description}

To select the tasks and datasets suitable for our objective, we analysed the corpora from the most recent survey on multilingual datasets \cite{yu2022beyond} that keeps real-time track of the multilingual data. We selected the datasets for the similar tasks of classification and sentiment analysis, and filtered out the datasets that contain multilingual data for short inputs - lemmas, word-pairs, sentences, tweets and short statements. We then manually analysed each dataset based on how the multilingual data was collected, and filtered out parallel multilingual datasets. Finally, we filtered out the datasets that focus on specific narrow domains. This method narrowed down the scope of our interest to the dataset that was created recently as part of \emph{SemEval-2023 Task 3: ``Detecting the genre, the framing, and the persuasion techniques in online news in a multi-lingual setup''} \cite{semeval2023task3}.

Prior to SemEval 2023, a number of other related challenging multilingual misinformation and propaganda detection tasks were addressed in SemEval (\url{https://semeval.github.io}) shared tasks, including detection of hyperpartisan content \cite{kiesel-etal-2019-semeval}, sarcasm \cite{abu-farha-etal-2022-semeval}, and a smaller set of persuasion techniques in textual \cite{da-san-martino-etal-2020-semeval} and multimodal \cite{dimitrov-etal-2021-semeval} data. 
The Shared Task 3 within SemEval 2023 challenge extended this prior work on persuasion techniques by introducing new kinds of persuasion techniques, as well as addressing two other related sub-tasks, namely news genre categorisation and framing detection. 

\paragraph{Sub-task 1: News Genre Categorisation.}
Given a news article, determine whether it is objective news reporting, an opinion piece, or satire.

\paragraph{Sub-task 2: Framing Detection.}
Given a news article, identify one or more of fourteen framing dimensions used: \emph{Economic}, \emph{Capacity and resources}, \emph{Morality}, \emph{Fairness and equality}, \emph{Legality, constitutionality and jurisprudence}, \emph{Policy prescription and evaluation}, \emph{Crime and punishment}, \emph{Security and defense}, \emph{Health and safety}, \emph{Quality of life}, \emph{Cultural identity}, \emph{Public opinion}, \emph{Political}, \emph{External regulation and reputation}.
The set of framing techniques used in this shared task was defined following a pre-existing taxonomy \cite{card-etal-2015-media}.

\paragraph{Sub-task 3: Persuasion Techniques Detection.}
Given a paragraph of a news article, identify zero or more out of 23 persuasion techniques used (see \nameref{S1_Appendix_ST3cats} for a detailed list of the techniques). The set of techniques represents an extension of the taxonomy used in previous SemEval datasets \cite{da-san-martino-etal-2019-fine, dimitrov-etal-2021-semeval}. The task additionally provides 6 high-level categories that subsume similar persuasion techniques. Although the task is paragraph level, each of the articles has at least one labelled paragraph.

It should be noted that three of the systems that participated in the original SemEval-2023 Task 3 \cite{semeval2023task3} evaluation exercise used adapters. Teams HHU \cite{HHUSemeval2023task3} and NAP \cite{NAPSemeval2023task3} entered only sub-task 3, in which they used adapters, whereas SheffieldVeraAI \cite{wu2023sheffieldveraai} applied adapters to sub-tasks 1 and 2. Initial performance analysis in these sub-tasks showed the effect of adapters to be inconsistent across the different sub-tasks. Namely, adapters achieved higher average performance for monolingual models in sub-task 1, while hindering performance of monolingual models in sub-task 2 but achieving better results there for multilingual models. 
This evidence provides an even stronger motivation gaining better understanding of the effectiveness of adapter methods across a range of classification tasks ranging in difficulty.

These three sub-tasks use broadly overlapping data, i.e. same input articles, however, differ in properties and the summary statistics for the datasets, which are summarised in Table \ref{tab:dataset-statistics}.

\begin{table}[H]
\begin{adjustwidth}{-2.25in}{0in}
\centering
\caption{\bf{Properties of genre, framing and persuasion technique detection tasks}}
\begin{tabular}{|l|c|c|c|}
\hline
Comparison criteria & Sub-task 1 & Sub-task 2 & Sub-task 3 \\ \hline
Task type & Genre Detection & Framing Detection & Persuasion Techniques \\ \hline
Input Type & Whole document & Whole document & Paragraphs \\ \hline
Granularity & Multiclass & Multilabel & Multilabel \\ \hline
Official scoring metric & \Fmacro & \Fmicro & \Fmicro \\ \hline
Number of classes & 3 & 14 & 23 \\ \hline
Avg number of tokens &  1,157 & 1,157 & 74 \\ \hline
Train set size & 1,234 & 1,238 & 10,927 \\ \hline
Number of source languages & 6 & 6 & 6 \\ \hline
Number of target languages & 9 & 9 & 9 \\ \hline
Task subjectivity & High & Low & Medium \\ \hline
Data imbalance & 12.7 & 4.6 & 44 \\ \hline
\end{tabular}
\label{tab:dataset-statistics}
\end{adjustwidth}
\end{table}

We estimate the data imbalance as a ratio of the class samples for the most frequent class in the training set to that for the most rare one. Below, we discuss some of the individual properties in more detail.

\paragraph{Class Imbalance:} 
In addition to posing multilingual and multiclass classification challenges, the data for these three sub-tasks is highly imbalanced, which adds further difficulty. In particular, the class distribution for sub-task 1 is highly skewed, with 76\% falling into the {\em opinion} class and {\em satire} accounting for only less than 6\% of the data. For sub-task 2, the distribution of classes is also uneven, but is less skewed compared to sub-task 1. The most common frame is {\em Political} which appears in 49.4\% of the training articles. The least common frame is {\em Cultural Identity} which appears in just 10.8\% of the articles. Finally, in sub-task 3 {\em loaded language}, {\em doubt} and {\em name calling} are the most common persuasion techniques, accounting for 22\%, 15.6\% and 12.8\% of the training paragraphs, respectively. The remaining 20 classes, on average, account for 2.5\% of the training set, totalling 49.6\% together. Particularly, {\em appeal to time}, {\em whataboutism} and {\em red herring} are the least frequent persuasion techniques, representing 0.5\%, 0.5\% and 0.7\% of the training paragraphs, respectively.

\paragraph{Dataset statistics: multilinguality, size and input length:} Three sets of data are provided for each language and task: labelled \emph{training} and \emph{development} (except for unseen languages), and  unlabelled \emph{testing}.

The task organisers provided test data in nine languages: English (EN), French (FR), German (DE), Georgian (KA), Greek (EL), Italian (IT), Polish (PL), Russian (RU), and Spanish (ES). Three of the languages (Georgian, Greek and Spanish) are `surprise' languages, meaning that no corresponding labelled training data exists in the dataset. Therefore, in order to make predictions for these languages, their test set must either be translated to a `seen' language, or a multi-lingual approach capable of supporting zero-shot evaluation must be applied. For the remaining 6 languages, labelled training and validation data is included in the dataset. 

It must be noted that the task organisers have not yet released the gold labels for the test set in order to prevent researchers from overfitting their systems. This means that detailed error analysis can only be carried out on the six languages for which the development sets are available.  

Tables~\ref{tab:data_st1}, \ref{tab:data_st2} and \ref{tab:data_st3} show the detailed analysis of the training, development and test data used in sub-tasks 1, 2 and 3 respectively. The average length, calculated in the number of tokens, was estimated using the tokenizer for RoBERTa-large  model \cite{liu2019roberta}, since this is the model we use in our experiments. For the training and development sets in sub-task 3, the average length was calculated for the paragraphs that have at least one persuasion technique assigned. 
For the test sets in sub-task 3, the average length includes every paragraph due to the lack of gold-standard labels for the test data. This is why the number of examples in the test-sets for sub-task 3 is significantly higher than that for the training and development sets. However, not all of these examples are expected to contain at least one persuasion technique. 

\begin{table}[H]
\caption{\textbf{Data statistics per language for sub-task 1: Genre Detection.}}
    \begin{tabular}{|l|c|c|c|c|c|c|}  
    \hline 
        \multirow{2}{*}{Language} & \multicolumn{3}{c|}{Number of examples} & \multicolumn{3}{c|}{Average number of tokens} \\ \cline{2-7}
         & \multicolumn{1}{l|}{Training} & \multicolumn{1}{l|}{Development} & \multicolumn{1}{l|}{Test} & \multicolumn{1}{l|}{Training} & \multicolumn{1}{l|}{Development} & \multicolumn{1}{l|}{Test} \\ \hline
        EN & 433 & 83 & 54 & 1,307 & 1,066 & 978 \\ \hline
        FR & 158 & 54 & 50 & 1,241 & 1,025 & 927 \\ \hline
        DE & 132 & 45 & 50 & 995 & 913 & 1,203 \\ \hline
        IT & 226 & 77 & 61 & 975 & 856 & 958 \\ \hline
        PL & 144 & 50 & 47 & 1,354 & 1,438 & 1,935 \\ \hline
        RU & 142 & 49 & 72 & 1020 & 797 & 547 \\ \Cline{1pt}{1-7}
        ES & 0 & 0 & 30 & N/A & N/A & 838 \\ 
        \hline
        EL & 0 & 0 & 64 & N/A & N/A & 1,071\\ \hline
        KA & 0 & 0 & 29 & N/A & N/A & 429\\ \hline
    \end{tabular}
    \label{tab:data_st1}
\end{table}
    
\begin{table}[H]
    \caption{\textbf{Data statistics per language for sub-task 2: Framing Detection.}}
    \begin{tabular}{|l|c|c|c|c|c|c|}  
    \hline 
        \multirow{2}{*}{Language} & \multicolumn{3}{c|}{Number of examples} & \multicolumn{3}{c|}{Average number of tokens} \\ \cline{2-7}
         & \multicolumn{1}{l|}{Training} & \multicolumn{1}{l|}{Development} & \multicolumn{1}{l|}{Test} & \multicolumn{1}{l|}{Training} & \multicolumn{1}{l|}{Development} & \multicolumn{1}{l|}{Test} \\ \hline
        EN & 433 & 83 & 54 & 1,307 & 1,066 & 978 \\ \hline
        FR & 158 & 53 & 50 & 1,196 & 1,059 & 927 \\ \hline
        DE & 132 & 45 & 50 & 1,008 & 875 & 1,203 \\ \hline
        IT & 227 & 76 & 61 &  965 & 885 & 958 \\ \hline
        PL & 145 & 49 & 47 &  1,369& 1,397 &  1,935\\ \hline
        RU & 143 & 48 & 72 & 1,009 & 827 & 547 \\ \Cline{1pt}{1-7}
        ES & 0 & 0 & 30 & N/A & N/A & 838 \\ \hline
        EL & 0 & 0 & 64 & N/A & N/A & 1,071\\ \hline
        KA & 0 & 0 & 29 & N/A & N/A & 429 \\ \hline
    \end{tabular}
    \label{tab:data_st2}
\end{table}

\begin{table}[H]
    \caption{\textbf{Data statistics per language for sub-task 3: Persuasion Techniques.}}
    \begin{tabular}{|l|c|c|c|c|c|c|}  
    \hline 
        \multirow{2}{*}{Language} & \multicolumn{3}{c|}{Number of examples} & \multicolumn{3}{c|}{Average number of tokens} \\ \cline{2-7}
         & \multicolumn{1}{l|}{Training} & \multicolumn{1}{l|}{Development} & \multicolumn{1}{l|}{Test} & \multicolumn{1}{l|}{Training} & \multicolumn{1}{l|}{Development} & \multicolumn{1}{l|}{Test} \\ \hline
        EN & 3,610 & 1,103 & 11,466 & 88 & 37 & 65 \\ \hline
        FR & 1,693 & 437 & 7,140 & 97 & 108 & 91 \\ \hline
        DE & 1,251 & 405  & 11,060 & 95 & 87 & 76 \\ \hline
        IT & 1,742 & 594 & 8,302 & 99 & 91 & 99 \\ \hline
        PL & 1,228 & 415 & 14,084 & 109 & 122 & 90  \\ \hline
        RU & 1,232 & 310 & 8,414 & 85 & 80 & 66 \\ \Cline{1pt}{1-7}
        ES & 0 & 0 & 1,320 & N/A & N/A & 76 \\ \hline
        EL & 0 & 0 & 3,792 & N/A & N/A & 72 \\ \hline
        KA & 0 & 0 & 640 & N/A & N/A & 78 \\ \hline
    \end{tabular}
    \label{tab:data_st3}
\end{table}

Sub-tasks 1 and 2 use the same set of articles in the test set, while the accumulative set of articles used in the development and training sets is also identical for these two sub-tasks, their assignment to a certain set varies slightly. This is why, as we can see from Tables~~\ref{tab:data_st1} and \ref{tab:data_st2}, the data statistics for these sub-tasks are quite similar. As can be seen, the distribution of the training examples across the languages is not even, with EN accounting for almost 4 times as many articles as DE, RU and PL. We can also observe that the average length of the articles is highly dependent on the language, with articles in the test set for Georgian (KA) being more than 4.5 times shorter than that for the articles in the test set for Polish (PL). This aspect is important for our experiments since it suggests that the models are more likely to omit important information for certain languages compared to others, due the the limitation of transformer models in terms of the input length. Another observation is that the training set is not always representative of the test set. For example, the articles in the test set for Russian (RU) are twice shorter, on average, than the articles in the training set for this language. The difference in terms of the input length between training and test data is less significant for sub-task 3, which uses paragraphs as an input, and for which all the inputs are within the limit of transformer models.

We analysed the languages in the training and test sets in terms of the amount of resources. For each language, we considered the number of training examples in that language or in a language from the same language family and the amount of pre-training data in that language for the model that we use in our experiments. Our analysis places Georgian as a clear outlier and a low-resource language, having significantly less pre-training data than the other 8 languages and no training data in Georgian or a related language. English, on the other hand, is the highest-resource language, benefiting from most training and pre-training data and additional training data in a related language. The details of this analysis are provided in \nameref{S4_Appendix_Resourse}.

\paragraph{Task subjectivity:} As specified in the annotation instructions \cite{annot-propaganda}, the subjectivity differs across the sub-tasks. Sub-task 1 relies on a very nuanced analysis of the whole article and on commonsense knowledge, since, as mentioned by the organisers, the satirical articles often ``tend to mimic true articles" and to mention ``real-world individuals, organisations, and events", while the distinction between opinionated and objective reporting can lie in the certain ways the reporters tend to balance out the reported opinions. The authors also highlight that ``the borders between opinion and reporting might be sometimes blurred" and ``a news article which contains some small text fragment, e.g., a sentence that appears satirical" does not normally trigger a satire genre. The task of framing detection tends to be more reliant on a certain linguistic information as the annotation instructions ask the annotators to specify exact text spans corresponding to a certain frame. The authors additionally provide examples of the discussion topics that trigger certain frames. For example, ``costs, benefits, or other financial implications" usually are indicative of the ``economic" frame. Finally, sub-task 3 is the most fine-grained task as it provides span-level annotations of the propaganda techniques. The detection of persuasion techniques usually relies on a certain argumentative structure and linguistic triggers, such as the mention of an entity that is considered an authority (``appeal to authority" technique), associating an opponent with a group, event or concept that has negative connotations (``guilt by association" technique),  ``the noun phrase, the adjective that constitutes the label and/or the name" (``name calling or labeling" technique) or ``text fragments that repeat the same message or information that was introduced earlier" (``repetition" technique). More objective assessment of the subjectivity is not possible due to the lack of the inter-annotator agreement scores \cite{semeval2023task3}.
\paragraph{Dataset collection:} The data is extracted from both mainstream and alternative media sources, collected through news aggregators (e.g. Google News, Europe Media Monitor) and fact-checking organisations (e.g. MediaBiasFactCheck, NewsGuard), respectively. All of the news articles were published between 2020 to mid 2022. The text of each article was  extracted automatically from the HTML source of each web page by using either the text-gathering tool Trafilatura \cite{barbaresi-2021-trafilatura} or a site-specific procedure. Notably, this process is error prone as it sometimes includes textual content which is not a part of the news article itself, such as web polls, newsletter sign-up forms, and author information. For English, a pre-existing dataset was also utilised \cite{da-san-martino-etal-2019-fine}, but the organisers of the shared task did not make it sufficiently clear as to what other English data was included in the new dataset.



\section*{Methods}

\subsection*{Training scenarios}
While our primary focus is on the multilingual fine-tuning scenario, we introduce two additional settings where models are trained on English-only data in order to investigate whether the effectiveness of each training method differs depending on the composition and the size of the training set.

These three different training scenarios are summarised below:
\begin{itemize}
    \item \textbf{Multilingual Joint (many-to-many)}: models are fine-tuned using all training data in the original 6 languages.
    \item \textbf{English + Translations (one-to-many)}: models are fine-tuned on all the original English training data and English translations of the training data in the other 5 languages. It is important to mention that the test data in this scenario was kept in its original languages, meaning that the predictions on all the languages except for English were made in a zero-shot cross-lingual way.
    \item \textbf{English Only (one-to-many)}: models are fine-tuned on only the original English data in the training set. Similarly to the `English + Translations' scenario, the test set was not translated into English.
\end{itemize}

The choice of the three training scenarios above is motivated by the fact that we want to evaluate the effect of two different factors on each of the training techniques:
\begin{enumerate}
    \item The `English + Translations' training scenario enables the evaluation of the effect of multilinguality. In particular, we want to compare the effect of having multilingual training data against the scenario where training data is available in only one language. By translating other languages into English, we compose a dataset consisting of the same number of training examples but without having the language diversity. 
    This eliminates the possibility that differences in performance across the three methods could be due to the size of the training data for each language. At the same time, machine translation, as a specific transfer paradigm for cross-lingual learning~\cite{PIKULIAK2021113765}, may introduce some level of noise and thus break the required correspondence between the original and translated sample.
    
    \item The `English Only' training scenario enables the analysis of the effect of training data size on each method. This can be achieved by comparing performance on `English Only' training data against performance on `English + Translations' data, where the only difference between the two is in the number of training examples. This training scenario, however, is not directly comparable against the multilingual training scenario, since it does not eliminate the possibility that differences in the method's performance could be due to the different linguistic characteristics of the multilingual training data.
\end{enumerate}

\subsection*{Training techniques}

In selecting the PEFT methods for our further analysis, we performed a detailed ablation study on sub-task 1, where we compare the adapter, LoRA and BitFit methods in multilingual and cross-lingual classification scenarios (see \nameref{S2_Appendix_bitfit}). As can be seen, the BitFit method does not reach a performance comparable to that of LoRA and adaptor methods for any of languages in any of the classification scenarios. We therefore exclude this PEFT technique from our further experiments.

We experiment with XLM-RoBERTa Large \cite{conneau-etal-2020-unsupervised}, using the following training techniques:
\begin{itemize}
    \item \textbf{Full fine-tuning (FFT)}: All parameters of the model are updated during fine-tuning.
    \item \textbf{Low-Rank Adaptation (LoRA)}: The model's parameters are frozen and LoRA matrices (key, query, value) are added to both the MLP and attention layers.
    \item \textbf{Bottleneck Adapter (Adapter)}: The model's parameters are frozen and bottleneck adapters in the Pfeiffer configuration \cite{pfeiffer-etal-2020-mad} are added to all the layers. As we discuss further, the choice of the adapter configuration was suggested by our ablation experiments that demonstrated the Pfeiffer configuration to outperform the Houlsby version on average (\nameref{S3_Appendix_ROBERTA}). 
\end{itemize}

The next section describes the methodology behind each training technique and training scenario.

\subsection*{Experimental setup}

\paragraph{Model and hyperparameter selection: } 

When selecting the model size, we took into account prior comparison of the LoRA and adapter-based PEFT methods for different sizes of RoBERTa model across a wide range of tasks and found a clear preference for a larger model size \cite{pfeiffer-etal-2020-mad, hu2021lora}. For both FFT and PEFT methods, we additionally conducted  the comparison of XLM-RoBERTa Base with XLM-RoBERTa Large on our sub-tasks in the default training scenario which includes all the data for each sub-task in the original form (the `Multilingual Joint' scenario). The results of this comparison are presented in \nameref{S3_Appendix_ROBERTA}. As can be observed, a large model demonstrates consistently better average performance across all three sub-tasks. Therefore, based on both prior evidence for PEFT techniques and our ablation studies for FFT, we resolved to using XLM-RoBERTa Large in our main experimental setup. 

Choosing between the Pfeiffer \cite{pfeiffer-etal-2020-adapterhub} and Houlsby \cite{houlsby2019parameter} configuration of the adapter method, we performed the comparison of both adapters for the `Multilingual Joint' training scenario. The results of this comparison are presented in \nameref{S5_Appendix_adapter}. As can be seen, Pfeiffer adapter shows slight average advantage over Houlsby one in all three sub-tasks. We therefore use the Pfeiffer configuration in the rest of our experiments.

For the best hyperparameters, we first perform a search for each training scenario and training technique within each sub-task. The search is performed on the original development set as provided by the organisers of SemEval 2023 Task 3. The best configuration obtained for each method can be found in Table \ref{tab:hyperparameters}. One needs to bear in mind that our objective is to maximise model performance per training scenario for each sub-task rather than to minimise the computational costs. In other words, greater parameter efficiency could be possible, but at the cost of model performance.

\begin{table}[H]
\centering
\caption{\bf{Hyperparameters.}}
\begin{tabular}{|lccc|}
\hline
\multicolumn{4}{|c|}{\textbf{Full fine-tuning (FFT)}} \\ \hline
\multicolumn{1}{|l|}{Hyperparameter} & \multicolumn{1}{c|}{Sub-task 1} & \multicolumn{1}{c|}{Sub-task 2} & Sub-task 3 \\ \hline
\multicolumn{1}{|l|}{Learning Rate} & \multicolumn{1}{c|}{1.00E-05} & \multicolumn{1}{c|}{3.00E-05} & 3.40E-04 \\ \hline
\multicolumn{1}{|l|}{Batch size} & \multicolumn{1}{c|}{16} & \multicolumn{1}{c|}{8} & 32 \\ \hline

\noalign{\vskip 4pt} \hline 

\multicolumn{4}{|c|}{\textbf{Low-Rank Adaptation (LoRA)}} \\ \hline
\multicolumn{1}{|l|}{Hyperparameter} & \multicolumn{1}{c|}{Sub-task 1} & \multicolumn{1}{c|}{Sub-task 2} & Sub-task 3 \\ \hline
\multicolumn{1}{|l|}{Learning rate} & \multicolumn{1}{c|}{7.00E-06} & \multicolumn{1}{c|}{3.00E-4} & 1.59E-03 \\ \hline
\multicolumn{1}{|l|}{Batch size} & \multicolumn{1}{c|}{16} & \multicolumn{1}{c|}{8} & 32 \\ \hline
\multicolumn{1}{|l|}{Rank} & \multicolumn{1}{c|}{8} & \multicolumn{1}{c|}{8} & 2 \\ \hline
\multicolumn{1}{|l|}{Layers} & \multicolumn{1}{c|}{all} & \multicolumn{1}{c|}{all} & all \\ \hline
\multicolumn{1}{|l|}{Dropout} & \multicolumn{1}{c|}{5.00E-01} & \multicolumn{1}{c|}{5.00E-01} & 5.00E-01 \\ \hline
\multicolumn{1}{|l|}{Attention matrix} & \multicolumn{1}{c|}{k,q,v} & \multicolumn{1}{c|}{k,q,v} & k,q,v \\ \hline

\noalign{\vskip 4pt} \hline 

\multicolumn{4}{|c|}
{\textbf{Bottleneck Adapter}} \\ \hline
\multicolumn{1}{|l|}{Hyperparameter} & \multicolumn{1}{c|}{Sub-task 1} & \multicolumn{1}{c|}{Sub-task 2} & Sub-task 3 \\ \hline
\multicolumn{1}{|l|}{Learning rate} & \multicolumn{1}{c|}{3.16E-05} & \multicolumn{1}{l|}{2.00E-4} & 4.30E-04 \\ \hline
\multicolumn{1}{|l|}{Batch size} & \multicolumn{1}{c|}{16} & \multicolumn{1}{c|}{8} & 32 \\ \hline
\multicolumn{1}{|l|}{Reduction factor} & \multicolumn{1}{c|}{4} & \multicolumn{1}{c|}{8} & 8 \\ \hline
\end{tabular}

\label{tab:hyperparameters}
\end{table}

For the bottleneck adapter, we use the default Pfeiffer configuration \cite{pfeiffer-etal-2020-adapterhub} by adding the adapter after each `ffn' sub-layer. Despite prior ablation studies showing that the lowest 4 layers have little contribution to the performance \cite{dettmers2023qlora}, our objective is to maximise our performance and to make the setting comparable with LoRA where we use all the layers.

\paragraph{Text preprocessing: }
As shown in Table \ref{tab:dataset-statistics}, sub-tasks 1 and 2 have an average number of tokens per article of 1,157. Thus in those sub-tasks, 80.0\% of articles are truncated to a maximum of 512 tokens. In contrast, sub-task 3 presents an average number of tokens per sequence of 74, thus all training sentences are fully encoded without loss of information. For sub-task 1, the articles that are longer than 512 tokens are separated into sentences, which are then sampled sequentially from the beginning and the end of the article, preserving the original order, until the maximum of 512 tokens is reached. Such a truncation approach is motivated by our experiments on sub-task 1 data during the competition stage of SemEval 2023 Task 3 \cite{wu2023sheffieldveraai}. This approach yielded a significant improvement in the \Fmacro  score  over the setting that simply truncates texts to the first 512 tokens. This improvement can potentially be explained by the fact that the instructions for human annotators in sub-task 1 highlighted the importance of opinionated sentences which tend to be found towards the end of the articles.

We perform text preprocessing for sub-tasks 1 and 2 by applying the following steps for all languages:
\begin{itemize}
    \item a full stop is added at the end of each title;
    \item duplicate sentences directly following each other are removed;
    \item the @ symbol is removed from any Twitter handles;
    \item hyperlinks to websites and images are also removed.
\end{itemize}
English articles were further preprocessed as follows:
\begin{itemize}
    \item text promoting sharing on different social media platforms was removed from the bottom of the articles;
    \item sentences encouraging user participation in online polls, comments, or advertisements were also removed;
    \item sentences stipulating the site's terms of use were removed;
    \item removal of sentences indicating licensing and containing phrases such as `reprinted with permission', `posted with permission' and `all rights reserved';
    \item sentences detailing author biographies were also removed.
\end{itemize}

For sub-task 3, preliminary experiments found no performance gains when text preprocessing was applied, thus our experiments for this sub-task use directly the original text. Importantly, for sub-task 3 experiments, we include sentences that do not have assigned labels into the training data by assigning them a vector of zeros to indicate that they do not belong to any class. This approach was shown to significantly improve classification performance on this sub-task in our initial experiments \cite{wu2023sheffieldveraai, hromadka2023kinitveraai}. The size of the training set displayed in Table~\ref{tab:dataset-statistics} (10,927 examples) is based on the number of labelled examples. When unlabelled sentences are added, training data for sub-task 3 grows to 20,704 instances.

The multilabel sub-tasks 2 and 3 use confidence thresholds of 50\% and 30\%, respectively, after applying a sigmoid activation function to the logits. The confidence threshold for sub-task 3 is purposefully lower and was selected according to our previous experiments \cite{hromadka2023kinitveraai}, which revealed that its careful calibration can significantly influence the performance of the model.

\paragraph{Training scenarios:} We experiment with the three training scenarios described previously. All models are trained on the original training split provided by the task organisers, using either data in all 6 seen languages; all EN data and the translations into English of the data in the other 5 languages; or only using the EN part of the training data. 
Each model is then evaluated on the task organiser's test split (6 seen and 3 surprise languages), without translation.

Aligned with previous research \cite{pires-etal-2019-multilingual, xenouleas2022realistic, he-etal-2021-effectiveness}, we define the ‘unseen’ language is a ‘target’ language on which a cross-lingual zero-shot prediction is done and which is different from the ‘source’ language on which the task-specific fine-tuning of the transformer model was performed. The three unseen test set languages - Greek, Georgian, and Spanish - allow us to evaluate the zero-shot cross-lingual transfer learning capabilities of the training techniques trained in the first, fully multilingual, setting. In the other two training scenarios (`English + Translations' and `English Only'), the remaining 8 languages (FR, DE, IT, PL, RU, ES, EL and KA) provide an insight into the models' performance in the cross-lingual zero-shot setting.

\paragraph{Evaluation metrics: } The performance of the different training techniques is then compared using two sets of criteria: (1) computational resource efficiency; (2) classification performance. For the latter, both \Fmicro and \Fmacro are reported as performance metrics for all three sub-tasks. However, it must be noted that the SemEval 2023 Task 3 organisers used only \Fmacro as the official scoring metric for sub-task 1, whereas sub-tasks 2 and 3 used only \Fmicro. Therefore, where a more detailed language-specific analysis is carried out in this paper, only the respective official metric for each sub-task is provided. 

Mean and standard deviation are computed over three different random seed initialisations. 

Resource efficiency is measured through four metrics: (i) the peak amount of VRAM used during training; (ii) speedup relative to the fully fine-tuned method, which is the number of training steps per second ${N_m} / {t_m}$ of the respective method (LoRA or Adapter) divided by the number of training steps per second of the fully fine-tuned method $N_{FFT} / {t_{FFT}}$ (Equation~\ref{eq:speedup});
(iii) the number of trainable parameters; and (iv) the number of non-trainable parameters.

\begin{equation}\label{eq:speedup}
    S^*=\frac{\frac{N_m}{t_m}}{\frac{N_{FFT}}{t_{FFT}}}
\end{equation}

\paragraph{Implementation details: } All experiments were performed with the AdapterHub framework \cite{pfeiffer-etal-2020-adapterhub}. 

In order to obtain the `English + Translations' data, we translate all available training and development data into English using Google Cloud Translation API.  The choice of the Machine Translation (MT) system was motivated by the recent extensive report on the evaluation of  MT systems \cite{savenkov-lopez-2022-state}. The results across all 11 language pairs and 9 domains analysed by the authors show that Google Translate is the state-of-the-art system based on the COMET score. We believe that this choice helps to reduce the potential noise caused by the translation, however, it is difficult to quantify the effect of the noise for the general case of news articles, since, as the  report  shows, the performance of the systems varies highly depending on the domain. Moreover, while we have the performance estimates for EN-DE, EN-FR, EN-ES and EN-IT pairs (which are comparable across all domains), we could not find  the relevant results for the remaining 4 language pairs (EN-PL, EN-KA, EN-RU, EN-EL). 

Our code is available on GitHub (\url{https://github.com/GateNLP/PEFT_FFT_multilingual}) and the Zenodo repository (\url{https://doi.org/10.5281/zenodo.10066649}).

\section*{Results}

The analysis of our results is structured around the following three main research questions:
\begin{description}
\item \textbf{RQ1:} How does the classification performance and computational costs of each training technique differ for each sub-task?
\item \textbf{RQ2:} How do the training scenarios (determining the diversity of the languages in the training set and its size) affect the performance of each training technique?
\item \textbf{RQ3:} How do the training techniques compare with each other for each training scenario and language?
\end{description}

The importance of the research question posed in this study is motivated by the limitations of prior research on the performance of PEFT techniques in the multilingual article-level classification tasks.

RQ1 provides a high-level analysis of each training technique under the best training scenario. This comparison is important because it provides the first to our knowledge examination of the PEFT and FFT techniques on non-parallel texts that are not restricted to a narrow domain and range in length from paragraphs to long articles. This RQ is particularly novel for the LoRA method since, as we highlighted above, no comparison of LoRA with adapter methods or FFT was previously performed for the multilingual multilabel classification task. We additionally provide the comparison of the computational costs as such a comparison was not previously performed for any of the sub-tasks in this study. It is not possible to assume that previous results would hold for our tasks given that the computational efficiency of PEFT methods can be sensitive to the input length \cite{hu2021lora}. Unlike previous analyses between bottleneck adaptors, LoRA and FFT that either perform a task-agnostic inference latency analysis or provide the number of trainable parameters for specific tasks \cite{hu2021lora}, our comparison focuses on more fine-grained statistics, such as the peak VRAM usage and the relative train time duration. The latter statistic is particularly important because the training process is typically more time-consuming than inference, and the relation between the number of trainable parameters and the training time is not linearly proportional.

RQ2 provides better insights into how the performance of each training technique changes in different training scenarios. As mentioned in the `Related work' section, previous experiments \cite{xenouleas2022realistic} conducted on non-parallel multilingual tests conclude that the zero-shot translation-based (translate-train and translate-test) approaches outperform cross-lingual ones for both FFT and adapter training techniques. However, their cross-lingual zero-shot approaches are limited to one-to-many scenarios, making it difficult to fairly compare the translation-based training settings to the joint multilingual scenarios due to the differences in the sizes of the training sets. We aim to address this gap by sequentially changing one aspect of the training process, by first eliminating the multilinguality of the training set whilst keeping the same size, and then reducing the size and eliminating potentially noisy translated texts from the training set. Additionally, previous joint multilingual experiments \cite{xenouleas2022realistic} do not leave any languages out to perform the zero-shot cross-lingual inference, which also makes it impossible to compare this approach to the monolingual ones. Finally, we add LoRA to our set of PEFT methods and provide novel insights on the performance of this technique under different scenarios in multilingual tasks.

RQ3 focuses on a different dimension of the problem and tries to answer which method is preferable depending on the amount and type of the training data. This analysis is motivated by the prior work by He et al. \cite{he-etal-2021-effectiveness} on short texts that concludes that adapter methods are particularly beneficial for low-resource and cross-lingual tasks. We therefore investigate how PEFT methods and FFT compare to each other as we reduce the number of training resources in certain languages and reduce the overall amount of training data.

\subsection*{Comparison of the computational and performance properties of training techniques}

To answer the first research question (RQ1), we select the best training scenario for each method and sub-task and compare the performance of the FFT model against the performance of the LoRA and adapter methods for each sub-task. The results of this comparison are reported in Table~\ref{tab:main-results}. We report mean scores after three runs with different random seeds along with standard deviations. The standard deviation is the square root of the average of the squared deviations from the mean.

\begin{table}[H]
\centering
\caption{\bf{Performance and computational costs for each sub-task and training technique.}}
\begin{tabular}{|llll|}
\hline
\multicolumn{4}{|c|}{Sub-task 1: Genre Detection} \\ \hline
\multicolumn{1}{|l|}{} & \multicolumn{1}{c|}{FFT} & \multicolumn{1}{c|}{LoRA} & \multicolumn{1}{c|}{Adapter} \\ \hline
\multicolumn{1}{|l|}{\Fmacro*} & \multicolumn{1}{c|}{\textbf{59.9 ± 3.1}} & \multicolumn{1}{c|}{57.9 ± 6.3} & \multicolumn{1}{c|}{58.0 ± 2.0} \\ \hline
\multicolumn{1}{|l|}{\Fmicro} & \multicolumn{1}{c|}{61.7 ± 7.5} & \multicolumn{1}{c|}{60.2 ± 3.9} &   \multicolumn{1}{c|}{\textbf{62.8  ± 5.7}}\\ \hline
\multicolumn{1}{|l|}{Best training scenario} & \multicolumn{1}{c|}{Multilingual Joint} & \multicolumn{1}{c|}{Multilingual Joint} &  Multilingual Joint\\ \hline
\multicolumn{1}{|l|}{Peak VRAM usage} & \multicolumn{1}{c|}{$\sim$39GB} & \multicolumn{1}{c|}{\textbf{$\sim$24GB}} & \multicolumn{1}{c|}{$\sim$28GB} \\ \hline
\multicolumn{1}{|l|}{Train Time relative to FFT} & \multicolumn{1}{c|}{1} & \multicolumn{1}{c|}{\textbf{0.59}} & \multicolumn{1}{c|}{0.67} \\ \hline
\multicolumn{1}{|l|}{Trainable parameters} & \multicolumn{1}{c|}{$\sim$560M} & \multicolumn{1}{c|}{$\sim$3.2M} & \multicolumn{1}{c|}{$\sim$26M} \\ \hline
\multicolumn{1}{|l|}{Non-trainable parameters} & \multicolumn{1}{c|}{0} & \multicolumn{1}{c|}{$\sim$560M} & \multicolumn{1}{c|}{$\sim$560M} \\ \hline

\noalign{\vskip 5pt} \hline 

\multicolumn{4}{|c|}{Sub-task 2: Framing Detection} \\ \hline
\multicolumn{1}{|l|}{} & \multicolumn{1}{c|}{FFT} & \multicolumn{1}{c|}{LoRA} & \multicolumn{1}{c|}{Adapter} \\ \hline
\multicolumn{1}{|l|}{\Fmacro} & \multicolumn{1}{c|}{\textbf{49.2 ± 7.4}} & \multicolumn{1}{c|}{45.3 ± 7.1} & \multicolumn{1}{c|}{47.1 ± 7.9} \\ \hline
\multicolumn{1}{|l|}{\Fmicro*} & \multicolumn{1}{c|}{\textbf{56.7 ± 6.1}} & \multicolumn{1}{c|}{53.4 ± 6.0} & \multicolumn{1}{c|}{54.8 ± 7.1} \\ \hline
\multicolumn{1}{|l|}{Best training scenario} & \multicolumn{1}{c|}{Multilingual Joint} & \multicolumn{1}{c|}{Multilingual Joint} & \multicolumn{1}{c|}{Multilingual Joint} \\ \hline
\multicolumn{1}{|l|}{Peak VRAM usage} & \multicolumn{1}{c|}{$\sim$23GB} & \multicolumn{1}{c|}{$\sim$18GB} & \multicolumn{1}{c|}{\textbf{$\sim$14GB}} \\ \hline
\multicolumn{1}{|l|}{Train Time relative to FFT} & \multicolumn{1}{c|}{1} & \multicolumn{1}{c|}{0.68} &  \multicolumn{1}{c|}{\textbf{0.56}} \\ \hline
\multicolumn{1}{|l|}{Trainable parameters} & \multicolumn{1}{c|}{$\sim$560M} & \multicolumn{1}{c|}{$\sim$4M} & \multicolumn{1}{c|}{$\sim$7M} \\ \hline
\multicolumn{1}{|l|}{Non-trainable parameters} & \multicolumn{1}{c|}{0} & \multicolumn{1}{c|}{$\sim$560M} & \multicolumn{1}{c|} {$\sim$560M} \\ \hline

\noalign{\vskip 5pt} \hline 

\multicolumn{4}{|c|}{Sub-task 3: Persuasion Techniques} \\ \hline
\multicolumn{1}{|c|}{} & \multicolumn{1}{c|}{FFT} & \multicolumn{1}{c|}{LoRA} & \multicolumn{1}{c|}{Adapter} \\ \hline
\multicolumn{1}{|l|}{\Fmacro} & \multicolumn{1}{c|}{\textbf{23.7 ± 5.0}} & \multicolumn{1}{c|}{\textbf{23.7 ± 6.9}} & \multicolumn{1}{c|}{20.6 ± 6.3} \\ \hline
\multicolumn{1}{|l|}{\Fmicro*} & \multicolumn{1}{c|}{41.8 ± 8.6} & \multicolumn{1}{c|}{\textbf{42.9 ± 9.5}} & \multicolumn{1}{c|}{42.2 ± 9.5} \\ \hline
\multicolumn{1}{|l|}{Best training scenario} & \multicolumn{1}{c|}{Multilingual Joint} & \multicolumn{1}{c|}{Multilingual Joint} & \multicolumn{1}{c|}{Multilingual Joint} \\ \hline
\multicolumn{1}{|l|}{Peak VRAM usage} & \multicolumn{1}{c|}{$\sim$20GB} & \multicolumn{1}{c|}{\textbf{$\sim$13GB}} & \multicolumn{1}{c|}{$\sim$16GB} \\ \hline
\multicolumn{1}{|l|}{Train Time relative to FFT} & \multicolumn{1}{c|}{1} & \multicolumn{1}{c|}{\textbf{0.56}} & \multicolumn{1}{c|}{0.71} \\ \hline
\multicolumn{1}{|l|}{Trainable parameters} & \multicolumn{1}{c|}{$\sim$560M} & \multicolumn{1}{c|}{$\sim$2M} & \multicolumn{1}{c|}{$\sim$7M} \\ \hline
\multicolumn{1}{|l|}{Non-trainable parameters} & \multicolumn{1}{c|}{0} & \multicolumn{1}{c|}{$\sim$560M} & \multicolumn{1}{c|}{$\sim$560M} \\ \hline
\end{tabular}
\begin{flushleft}
    The best scores and performance metrics appear in \textbf{bold}. The main metric for a certain sub-task is marked with an asterisk ($*$).
\end{flushleft}
\label{tab:main-results}
\end{table}

The results demonstrate that:

(1) \textbf{FFT and adapters perform better in sub-tasks 1 and 2, while LoRA performs better for sub-task 3}. We observe that for longer texts, such as the articles analysed in sub-tasks 1 and 2, FFT and adapter-based classification demonstrates better results on average than LoRA. At the same time, LoRA on average outperforms FFT and adapters for sub-task 3, which is trained on shorter texts.


(2) \textbf{`Multilingual Joint' training scenario performs best, regardless of the sub-task and training technique}. We observe a pattern of `Multilingual Joint` training scenario achieving the best results for all three sub-tasks as well as all three training techniques. This implies that, in general, training models on larger datasets with a variety of languages,
can be beneficial for both FFT and PEFT methods applied to the tasks with various properties. This effect has not been, however, consistently observed for all combinations of training scenarios and training techniques (the more detailed analysis per individual training scenarios is provided in the following section).

(3) \textbf{LoRA and adapters can save computational costs significantly}. By design, the PEFTs reduce the number of trainable parameters significantly (between 140 and 280 times less parameters). As a result, for sub-task 1 and 3, the utilisation of LoRA led to a significant decrease in the memory consumption --- from 39GB to 24GB (38\%), and from 20GB to 13GB (35\%) respectively. For sub-task 2, the adapter achieved the best memory efficiency, while decreasing the peak VRAM usage from 23GB to 14GB (39\%). The similar pattern can be observed for a total training time, which decreased to 56-71\% of the FFT training time.

(4) \textbf{Saving computational costs results in lower performance, some exceptions, however, exist}. Saving the VRAM usage and shortening training time is naturally reflected in the lower performance compared to that of FFT. Adapters are consistently outperformed by FFT in all three sub-tasks. However, in the case of sub-task 3, LoRA not only achieved highly comparable results, but also outperformed FFT for the most of languages (the difference is, however, not statistically significant --- the more detailed analysis is provided in the next sections). For sub-tasks 2 and 3, we also observe a higher standard deviation of the results, implying a higher instability of fine-tuning when PEFTs are applied.

\subsection*{Comparison of the effect of a training scenario on each training technique}

To answer the second question (RQ2), we compare FFT, LoRA and adapter training techniques across the three training scenarios introduced above, namely `Multilingual Joint', `English + Translations' and `English Only'. It should be noted that in the latter two scenarios all languages in the test set are unseen (except English) as the model did not have access to training data in those languages. The results are measured in the official sub-task metrics (\fmacro for sub-task 1 and \fmicro for sub-tasks 2 and 3) and are shown in Tables~\ref{detailed_st1}, \ref{detailed_st2} and \ref{detailed_st3}.

\begin{table}[H]
\small
{\begin{adjustwidth}{-2.25in}{0in}
\caption{\bf Sub-task 1: Genre Detection - Mean ± 1 STD \Fmacro scores.}
\def\arraystretch{1.1}
\begin{tabular}{|l|ccc|ccc|ccc|}
\hline
\multirow{2}{*}{Language} & \multicolumn{3}{c|}{Multilingual Joint} & \multicolumn{3}{c|}{English + Translations} & \multicolumn{3}{c|}{English Only} \\ \cline{2-10}
 & \multicolumn{1}{c|}{FFT} & \multicolumn{1}{c|}{LoRA} & Adapter & \multicolumn{1}{c|}{FFT} & \multicolumn{1}{c|}{LoRA} & Adapter & \multicolumn{1}{c|}{FFT} & \multicolumn{1}{c|}{LoRA} & Adapter \\ \hline

\noalign{\vskip 4pt} \hline 
 
EN & \multicolumn{1}{c|}{{52.7±0.5}} & \multicolumn{1}{c|}{49.4±0.4} & \multicolumn{1}{c|}{\textbf{52.8±0.2}} & \multicolumn{1}{c|}{53.1±1.1} & \multicolumn{1}{c|}{52.4±0.6} & \textbf{*53.5±0.9} & \multicolumn{1}{c|}{40.9±1.3} & \multicolumn{1}{c|}{39.1±1.2} & \textbf{42.2±0.6} \\ \hline
FR & \multicolumn{1}{c|}{\textbf{*69.7±1.2}} & \multicolumn{1}{c|}{67.4±2.3} & 67.5±0.9 & \multicolumn{1}{c|}{68.0±1.9} & \multicolumn{1}{c|}{\textbf{69.2±1.5}} & {68.4±0.7} & \multicolumn{1}{c|}{65.0±2.2} & \multicolumn{1}{c|}{66.3±0.5} & \textbf{66.7±3.7} \\ \hline
DE & \multicolumn{1}{c|}{66.3±0.5} & \multicolumn{1}{c|}{64.8±1.2} & \textbf{*67.2±0.8} & \multicolumn{1}{c|}{65.4±2.5} & \multicolumn{1}{c|}{63.9±3.0} &  \textbf{65.8±1.2}& \multicolumn{1}{c|}{62.1±4.2} & \multicolumn{1}{c|}{\textbf{64.2±2.8}} &  63.6±0.7\\ \hline
IT & \multicolumn{1}{c|}{52.2±1.4} & \multicolumn{1}{c|}{\textbf{*53.4±1.8}} &  52.0±3.1& \multicolumn{1}{c|}{52.1±1.7} & \multicolumn{1}{c|}{52.0±1.5} & \textbf{52.9±0.8} & \multicolumn{1}{c|}{45.1±3.6} & \multicolumn{1}{c|}{\textbf{47.3±1.8}} & 44.2±1.1 \\ \hline
PL & \multicolumn{1}{c|}{\textbf{*69.2±1.1}} & \multicolumn{1}{c|}{66.8±0.4} & 65.2±1.5 & \multicolumn{1}{c|}{61.3±0.7} & \multicolumn{1}{c|}{\textbf{64.0±0.7}} & 61.8±2.4 & \multicolumn{1}{c|}{58.9±2.2} & \multicolumn{1}{c|}{\textbf{60.6±1.4}} & 59.6±3.0 \\ \hline
RU & \multicolumn{1}{c|}{\textbf{*57.4±0.6}} & \multicolumn{1}{c|}{55.7±1.7} & 52.8±0.9 & \multicolumn{1}{c|}{52.4±1.3} & \multicolumn{1}{c|}{54.2±0.8} & \textbf{54.9±2.5} & \multicolumn{1}{c|}{\textbf{49.7±0.9}} & \multicolumn{1}{c|}{49.4±2.4} & 48.7±0.6 \\ \Cline{1pt}{1-10}

Average  & \multicolumn{1}{c|}{\textbf{*61.3±2.7}} & \multicolumn{1}{c|}{59.6±1.9} & 59.6±3.1 & \multicolumn{1}{c|}{58.7±1.8} & \multicolumn{1}{c|}{59.3±5.2} & \textbf{59.6±2.7} & \multicolumn{1}{c|}{53.6±3.5} & \multicolumn{1}{c|}{\textbf{54.5±6.1}} & 54.2±2.9\\ \hline
\noalign{\vskip 4pt} \hline 

ES & \multicolumn{1}{c|}{\textbf{*47.1±1.4}} & \multicolumn{1}{c|}{41.8±0.5} & 44.2±0.7 & \multicolumn{1}{c|}{44.5±2.7} & \multicolumn{1}{c|}{46.0±1.3} & \textbf{46.2±5} & \multicolumn{1}{c|}{\textbf{42.3±2.9}} & \multicolumn{1}{c|}{41.7±0.8} & 40.9±1.1\\ \hline
EL & \multicolumn{1}{c|}{40.8±2.4} & \multicolumn{1}{c|}{\textbf{41.4±2.7}} & 40.9±1.7 & \multicolumn{1}{c|}{\textbf{*43.5±2.1}} & \multicolumn{1}{c|}{42.9±1.7} & 42.2±3.6 & \multicolumn{1}{c|}{\textbf{38.9±2.7}} & \multicolumn{1}{c|}{38.6±1.7} & 37.5±0.8 \\ \hline
KA & \multicolumn{1}{c|}{\textbf{*83.3±2.1}} & \multicolumn{1}{c|}{80.8±5.0} & 79.2±1.8 & \multicolumn{1}{c|}{\textbf{81.0±3.2}} & \multicolumn{1}{c|}{79.6±4.1} & 77.5±2.2 & \multicolumn{1}{c|}{\textbf{76.7±3.3}} & \multicolumn{1}{c|}{72.9±1.5} & 74.8±2.4 \\ \Cline{1pt}{1-10}

Average & \multicolumn{1}{c|}{\textbf{*57.1±2.3}} & \multicolumn{1}{c|}{54.7±2.4} & 54.8±1.8 & \multicolumn{1}{c|}{\textbf{56.3±2.4}} & \multicolumn{1}{c|}{56.2±3.3} & 55.3±1.9 & \multicolumn{1}{c|}{\textbf{52.6±2.9}} & \multicolumn{1}{c|}{51.1±4.0} &  51.1±3.3\\ \hline
\noalign{\vskip 4pt} \hline 

\noalign{\vskip 4pt} \hline 

All & \multicolumn{1}{c|}{\textbf{*59.9±3.1}} & \multicolumn{1}{c|}{57.9±6.3} & 58.0±2.0 & \multicolumn{1}{c|}{57.9±3.4} & \multicolumn{1}{c|}{\textbf{58.2±3.8}} & 58.1±4.1 & \multicolumn{1}{c|}{\textbf{53.3±5.2}} & \multicolumn{1}{c|}{53.3±2.6} &  53.1±3.6 \\ \hline
\end{tabular}
\begin{flushleft}
    Best scores by language are marked with an asterisk ($*$). Best scores by training method for each training scenario are in \textbf{bold}.
\end{flushleft}
\label{detailed_st1}
\end{adjustwidth}}
\end{table}

\begin{table}[H]
\small
{\begin{adjustwidth}{-2.25in}{0in}
\caption{\bf{Sub-task 2: Framing Detection - Mean ± 1 STD \Fmicro scores.}}
\begin{tabular}{|l|ccc|ccc|ccc|}
\hline
\multirow{2}{*}{Language} & \multicolumn{3}{c|}{Multilingual Joint} & \multicolumn{3}{c|}{English + Translations} & \multicolumn{3}{c|}{English Only} \\ \cline{2-10}
 & \multicolumn{1}{c|}{FFT} & \multicolumn{1}{c|}{LoRA} & Adapter & \multicolumn{1}{c|}{FFT} & \multicolumn{1}{c|}{LoRA} & Adapter & \multicolumn{1}{c|}{FFT} & \multicolumn{1}{c|}{LoRA} & Adapter \\ \hline

\noalign{\vskip 4pt} \hline 
 
EN & \multicolumn{1}{c|}{\textbf{*55.8±0.2}} & \multicolumn{1}{c|}{52.2±1.7} & 55.7±2.0 & \multicolumn{1}{c|}{54.9±1.9} & \multicolumn{1}{c|}{54.3±1.7} & \textbf{55.1±1.5} & \multicolumn{1}{c|}{46.5±1.3} & \multicolumn{1}{c|}{46.2±1.3} & \textbf{47.8±0.7} \\ \hline
FR & \multicolumn{1}{c|}{\textbf{53.3±3.3}} & \multicolumn{1}{c|}{47.3±1.5} & 50.8±3.6 & \multicolumn{1}{c|}{\textbf{*54.5±2.5}} & \multicolumn{1}{c|}{52.9±2.5} & 50.3±2.5 & \multicolumn{1}{c|}{\textbf{44.2±1.6}} & \multicolumn{1}{c|}{41.0±0.8} & 41.7±0.4 \\ \hline
DE & \multicolumn{1}{c|}{63.1±1.9} & \multicolumn{1}{c|}{62.3±2.2} & \textbf{*64.2±1.0} & \multicolumn{1}{c|}{58.9±1.6} & \multicolumn{1}{c|}{55.8±1.8} & \textbf{59.1±3.3} & \multicolumn{1}{c|}{46.6±3.3} & \multicolumn{1}{c|}{45.7±1.1} & \textbf{49.1±2.9} \\ \hline
IT & \multicolumn{1}{c|}{\textbf{*59.9±1.9}} & \multicolumn{1}{c|}{56.8±1.6} & 58.2±1.0 & \multicolumn{1}{c|}{57.0±1.9} & \multicolumn{1}{c|}{55.7±0.4} & \textbf{59.5±1.0} & \multicolumn{1}{c|}{54.9±1.9} & \multicolumn{1}{c|}{\textbf{55.3±2.0}} & \textbf{55.3±0.6} \\ \hline
PL & \multicolumn{1}{c|}{\textbf{*65.2±0.8}} & \multicolumn{1}{c|}{61.0±0.8} & 64.1±1.7 & \multicolumn{1}{c|}{55.3±3.8} & \multicolumn{1}{c|}{54.1±1.5} & \textbf{56.3±1.9} & \multicolumn{1}{c|}{45.5±0.5} & \multicolumn{1}{c|}{\textbf{46.3±0.1}} & 46.1±2.7 \\ \hline
RU & \multicolumn{1}{c|}{\textbf{45.3±3.0}} & \multicolumn{1}{c|}{43.6±0.6} & 41.7±2.0 & \multicolumn{1}{c|}{40.3±2.0} & \multicolumn{1}{c|}{38.2±1.2} & \textbf{41.0±2.7} & \multicolumn{1}{c|}{\textbf{35.1±1.1}} & \multicolumn{1}{c|}{31.1±1.6} & 31.7±1.3 \\ \Cline{1pt}{1-10}

Average & \multicolumn{1}{c|}{\textbf{*57.1±7.3}} & \multicolumn{1}{c|}{53.9±7.5} & 55.8±8.6 & \multicolumn{1}{c|}{53.5±6.7} & \multicolumn{1}{c|}{51.8±6.8} & \textbf{53.6±7.0} & \multicolumn{1}{c|}{\textbf{45.5±6.3}} & \multicolumn{1}{c|}{44.3±7.9} & 45.3±8.0\\ \hline

\noalign{\vskip 4pt} \hline 

ES & \multicolumn{1}{c|}{\textbf{*52.7±2.1}} & \multicolumn{1}{c|}{51.7±3.0} & 49.1±2.0 & \multicolumn{1}{c|}{49.7±0.7} & \multicolumn{1}{c|}{\textbf{49.9±1.0}} & 46.6±2.5 & \multicolumn{1}{c|}{\textbf{41.8±1.9}} & \multicolumn{1}{c|}{35.9±3.8} & 36.4±2.1 \\ \hline
EL & \multicolumn{1}{c|}{\textbf{*54.9±1.7}} & \multicolumn{1}{c|}{51.4±1.2} & 54.1±2.9 & \multicolumn{1}{c|}{53.2±0.1} & \multicolumn{1}{c|}{50.7±0.7} & \textbf{55.5±1.1} & \multicolumn{1}{c|}{\textbf{47.2±1.2}} & \multicolumn{1}{c|}{43.6±0.2} & 47.1±1.5 \\ \hline
KA & \multicolumn{1}{c|}{\textbf{*60.1±4.2}} & \multicolumn{1}{c|}{53.9±4.8} & 55.3±1.6 & \multicolumn{1}{c|}{55.3±4.9} & \multicolumn{1}{c|}{52.8±4.0} & \textbf{57.0±4.6} & \multicolumn{1}{c|}{\textbf{51.4±0.9}} & \multicolumn{1}{c|}{45.3±1.7} & 50.1±2.1 \\ \Cline{1pt}{1-10}
Average & \multicolumn{1}{c|}{\textbf{*55.9±3.8}} & \multicolumn{1}{c|}{52.3±1.4} & 52.8±3.3 & \multicolumn{1}{c|}{52.7±2.8} & \multicolumn{1}{c|}{51.1±1.5} & \textbf{53.1±5.6} & \multicolumn{1}{c|}{\textbf{46.8±4.8}} & \multicolumn{1}{c|}{41.6±5.0} &  44.5±7.2\\ \hline

\noalign{\vskip 4pt} \hline 

All & \multicolumn{1}{c|}{\textbf{*56.7±6.1}} & \multicolumn{1}{c|}{53.4±6.0} & 54.8±7.1 & \multicolumn{1}{c|}{53.2±5.5} & \multicolumn{1}{c|}{51.6±5.4} & \textbf{53.4±6.2} & \multicolumn{1}{c|}{\textbf{45.9±5.6}} & \multicolumn{1}{c|}{43.4±6.9} & 45.0±7.3 \\ \hline
\end{tabular}
\begin{flushleft}
    Best scores by language are marked with an asterisk ($*$). Best scores by training method for each training scenario are in \textbf{bold}.
\end{flushleft}
\label{detailed_st2}
\end{adjustwidth}}
\end{table}

\begin{table}[H]
\small
{\begin{adjustwidth}{-2.25in}{0in}
\caption{\bf Sub-task 3: Persuasion Techniques - Mean ± 1 STD \Fmicro scores.}
\begin{tabular}{|l|ccc|ccc|ccc|}
\hline
\multirow{2}{*}{Language} & \multicolumn{3}{c|}{Multilingual Joint} & \multicolumn{3}{c|}{English + Translations} & \multicolumn{3}{c|}{English Only} \\ \cline{2-10}
 & \multicolumn{1}{c|}{FFT} & \multicolumn{1}{c|}{LoRA} & Adapter & \multicolumn{1}{c|}{FFT} & \multicolumn{1}{c|}{LoRA} & Adapter & \multicolumn{1}{c|}{FFT} & \multicolumn{1}{c|}{LoRA} & Adapter \\ \hline

\noalign{\vskip 4pt} \hline 
 
EN & \multicolumn{1}{c|}{34.9±1.7} & \multicolumn{1}{c|}{\textbf{37.7±0.9}} & 37.5±2.9 & \multicolumn{1}{c|}{42.2±0.8} & \multicolumn{1}{c|}{42.9±0.4} & \textbf{*44.0±1.0} & \multicolumn{1}{c|}{34.0±1.6} & \multicolumn{1}{c|}{\textbf{34.2±5.5}} & 33.3±3.4 \\ \hline
FR & \multicolumn{1}{c|}{45.9±1.2} & \multicolumn{1}{c|}{\textbf{*48.6±0.8}} & 45.7±1.9 & \multicolumn{1}{c|}{42.5±0.5} & \multicolumn{1}{c|}{41.9±1.8} & \textbf{42.9±1.1} & \multicolumn{1}{c|}{\textbf{28.9±2.5}} & \multicolumn{1}{c|}{28.3±6.0} & 24.2±4.3 \\ \hline
DE & \multicolumn{1}{c|}{52.1±1.9} & \multicolumn{1}{c|}{52.3±0.9} & \textbf{*53.0±0.7} & \multicolumn{1}{c|}{43.2±1.5} & \multicolumn{1}{c|}{\textbf{45.6±2.0}} & 44.0±1.9 & \multicolumn{1}{c|}{\textbf{29.1±2.5}} & \multicolumn{1}{c|}{27.4±4.2} & 26.7±6.9 \\ \hline
IT & \multicolumn{1}{c|}{55.1±2.5} & \multicolumn{1}{c|}{\textbf{*58.7±0.5}} & 58.1±1.6 & \multicolumn{1}{c|}{53.4±2.5} & \multicolumn{1}{c|}{52.9±0.9} & \textbf{54.0±1.5} & \multicolumn{1}{c|}{\textbf{33.6±3.0}} & \multicolumn{1}{c|}{33.0±7.5} & 32.6±3.0 \\ \hline
PL & \multicolumn{1}{c|}{40.4±3.2} & \multicolumn{1}{c|}{\textbf{*42.1±0.6}} & 41.9±2.7 & \multicolumn{1}{c|}{\textbf{37.8±1.7}} & \multicolumn{1}{c|}{36.9±1.6} & 37.5±1.2 & \multicolumn{1}{c|}{\textbf{23.4±2.2}} & \multicolumn{1}{c|}{22.3±1.9} & 19.6±5.2 \\ \hline
RU & \multicolumn{1}{c|}{40.9±1.4} & \multicolumn{1}{c|}{\textbf{*42.3±0.3}} & 39.2±2.6 & \multicolumn{1}{c|}{36.2±1.3} & \multicolumn{1}{c|}{35.4±0.8} & \textbf{37.1±0.4} & \multicolumn{1}{c|}{22.9±2.5} & \multicolumn{1}{c|}{\textbf{25.5±4.5}} & 17.9±3.7 \\ \Cline{1pt}{1-10}

Average & \multicolumn{1}{c|}{44.9±7.7} & \multicolumn{1}{c|}{\textbf{*46.9±7.7}} & \textbf{45.9±8.1} & \multicolumn{1}{c|}{42.5±6.0} & \multicolumn{1}{c|}{42.6±6.3} & \textbf{43.5±6.4} & \multicolumn{1}{c|}{\textbf{28.6±4.8}} & \multicolumn{1}{c|}{28.4±4.5} & 25.7±6.4 \\ \hline

\noalign{\vskip 4pt} \hline 

ES & \multicolumn{1}{c|}{37.7±2.3} & \multicolumn{1}{c|}{\textbf{*39.0±1.6}} & 36.7±1.3 & \multicolumn{1}{c|}{\textbf{38.3±0.9}} & \multicolumn{1}{c|}{35.3±1.5} & 37.8±0.8 & \multicolumn{1}{c|}{\textbf{26.1±1.2}} & \multicolumn{1}{c|}{26.0±4.7} & 23.0±5.1 \\ \hline
EL & \multicolumn{1}{c|}{26.7±0.9} & \multicolumn{1}{c|}{\textbf{*25.5±0.6}} & 25.4±1.5 & \multicolumn{1}{c|}{\textbf{22.6±0.1}} & \multicolumn{1}{c|}{21.8±0.9} & 21.9±0.9 & \multicolumn{1}{c|}{16.2±1.9} & \multicolumn{1}{c|}{\textbf{16.7±0.9}} & 12.5±1.2 \\ \hline
KA & \multicolumn{1}{c|}{\textbf{*42.6±1.0}} & \multicolumn{1}{c|}{40.4±3.1} & 42.2±2.4 & \multicolumn{1}{c|}{\textbf{36.8±3.1}} & \multicolumn{1}{c|}{35.4±1.6} & 36.7±1.8 & \multicolumn{1}{c|}{\textbf{29.4±4.2}} & \multicolumn{1}{c|}{24.2±6.1} & 22.4±8.7 \\ \Cline{1pt}{1-10} 
Average & \multicolumn{1}{c|}{\textbf{*35.7±8.1}} & \multicolumn{1}{c|}{34.9±8.2} & 34.7±8.5 & \multicolumn{1}{c|}{\textbf{32.6±8.7}} & \multicolumn{1}{c|}{30.8±7.8} & 32.1±8.9 & \multicolumn{1}{c|}{\textbf{23.9±6.9}} & \multicolumn{1}{c|}{22.3±4.9} & 19.3±5.9 \\ \hline

\noalign{\vskip 4pt} \hline 

All & \multicolumn{1}{c|}{41.8±8.6} & \multicolumn{1}{c|}{\textbf{*42.9±9.5}} & 42.2±9.5 & \multicolumn{1}{c|}{39.2±8.1} & \multicolumn{1}{c|}{38.7±8.7} & \multicolumn{1}{c|}{\textbf{39.7±8.8}} & \multicolumn{1}{c|}{\textbf{27.1±5.6}} & \multicolumn{1}{c|}{26.4±5.3} & 23.6±6.7 \\ \hline
\end{tabular}
\begin{flushleft}
    Best scores by language are marked with an asterisk ($*$). Best scores by training method for each training scenario are in \textbf{bold}.
\end{flushleft}
\label{detailed_st3}
\end{adjustwidth}}
\end{table}

(1) \textbf{The diversity of languages in the training set improves the average performance of the FFT technique.} For all three sub-tasks, we observe a significantly better  average classification performance when the training data is provided in the original 6 different languages as opposed to providing the same amount of data in English only. When looking at the individual languages, this effect holds for all languages in sub-task 1 except for English (the only seen language in the `English + Translations' training scenario) and Spanish (one of the unseen languages in the joint multilingual setting). The decreased performance on the English test set in the `Multilingual Joint' scenario for this task can potentially be explained with the `negative inference' effect highlighted in previous studies \cite{wang2020negative}. However, we do not observe this effect consistently across the sub-tasks. In particular, for sub-task 2, the only exception is French, which benefits from a monolingual training setting. Notably, the performance on the English test set decreases in the `English + Translations' scenario for sub-task 2, despite being trained on much more data in this language. Finally, for sub-task 3, French is the only language in the test set that benefits from being trained on monolingual English data, which is consistent with sub-task 2. 

(2) \textbf{The diversity of languages in the training set has an inconsistent effect on the performance of the LoRA training technique across the three sub-tasks.} While we observe an average decrease in classification performance in the `English + Translations' training scenario for sub-tasks 2 and 3, this setting improves the average results for sub-task 1. For sub-task 3, this effect holds for every language in the test set,  while for sub-task 2, LoRA improves performance on EN and FR when trained using monolingual English data. For sub-task 1, LoRA benefits from the multilingual training data when making predictions on 5 out of the 9 languages (DE, IT, PL, RU, and KA) and favours the monolingual English training setting for the 4 remaining languages (EN, FR, EL and ES). The difference is particularly high for Spanish, resulting in a slightly better average performance in a monolingual training scenario for sub-task 1.

(3) \textbf{The diversity of languages in the training set has an inconsistent effect on adapter classification performance across all tasks.} While the `English + Translations' training scenario insignificantly improves average performance in sub-task 1, it decreases the average performance in sub-tasks 2 and 3. Adapters benefit from a monolingual training scenario when making predictions on 6 out of the 9 languages (EN, FR, IT, RU, ES and EL) in sub-task 1. For sub-tasks 2 and 3, this is the case only for 3 languages (IT, EL and KA) and 2 languages (EN and ES) respectively.

(4) \textbf{Decrease in the size of the training set  decreases the performance of the FFT, LoRA as well as adapter method on all seen and zero-shot languages}. We observe a significant decrease in FFT's performance and both PEFTs in the `English Only' training scenario compared to the `English + Translation' setting across all the sub-tasks and for every language within each sub-task. This result is particularly noteworthy for the English test set as it indicates that even potentially noisy translated data is able to improve the performance on a given language as compared to using a smaller but better quality dataset.

In summary, the effect of removing language diversity from the training set is not consistent across SemEval 2023 sub-tasks and training techniques. Sub-task 1, in particular, demonstrates slight improvement in performance for LoRA and adapter methods when trained on `English + Translations' data.
All training techniques show a significantly decreased performance when trained on the original English-only data, demonstrating the importance of the size of the training data.

In the next section, we compare the results of the three training techniques within each of the three training scenarios.

\subsection*{Comparison of training techniques for each training scenario and language}

To answer the third research question (RQ3), we analyse whether the performance of FFT, LoRA and adapter methods is consistent across training scenarios or are certain scenarios more prone to the lack of training data and or language diversity within the training data. This analysis, on the top of results presented in Tables~\ref{detailed_st1}, \ref{detailed_st2} and \ref{detailed_st3} for sub-tasks 1, 2 and 3 respectively,  complement the previous RQ2.

(1) \textbf{FFT outperforms LoRA and adapter methods for sub-tasks 1 and 2 in a `Multilingual Joint` training scenario, while for sub-task 3, only zero-shot predictions on unseen languages benefit from the FFT method}. We observe that FFT yields best performance in a multilingual setting for most of the seen and unseen languages in sub-tasks 1 and 2. While the differences between FFT and LoRA are often less than 1\% for sub-task 1, sub-task 2 demonstrates a clear preference for the FFT classification approach. 

(2) \textbf{In an `English + Translations' training scenario, adapters outperform FFT across all sub-tasks and show better or on-par performance compared to LoRA.} A particularly clear preference for adapters can be observed for sub-task 2, where the majority of seen and unseen languages  benefit from this training technique. For sub-tasks 1 and 2, most of the zero-shot predictions on unseen languages favour FFT. The performance of the adapter method is particularly consistent on English, the only seen language in this training scenario, with all sub-tasks favouring the adapter classifier. 

(3) \textbf{In an `English Only' training scenario with less data, differences between FFT, LoRA and adapters become less obvious across all sub-tasks.} In this setting, the differences in the average classification performances between FFT, adapters and LoRA across all sub-tasks is often insignificant, with less than 1\% of one method over the other. For example, the difference between the adapter method and LoRA for sub-task 1 is 0.2\%, and the average performance of FFT and LoRA is the same and differs only in the confidence intervals. Similarly, for sub-task 2, FFT is comparable to the adapter method, and for sub-task 3, FFT is very close in the performance to LoRA. We observe that for sub-tasks 1 and 2, adapters perform better for EN (the only seen language), while for sub-task 3, LoRA yields better performance for the seen language (EN). 

(4) \textbf{The overall best performance across all training scenarios and training techniques is achieved in a `Multilingual Joint' training scenario.} While the FFT method works better for sub-tasks 1 and 2 in this setting, sub-task 3 shows a clear improvement when trained using the LoRA method. While sub-tasks 1 and 2 are consistent in favouring FFT for both seen and unseen (ES, EL, KA) languages, sub-task 3 favours LoRA when making the predictions on seen languages only. For unseen languages, sub-task 3 agrees with sub-tasks 1 and 2 in favouring the FFT classification approach.

Some of the languages demonstrate strong preferences towards certain training techniques and training scenarios:\\
(1) \textbf{English consistently favours adapter classification approach across all the training scenarios in sub-task 1}. 

(2) \textbf{FFT method yields best performance on Georgian zero-shot predictions across all settings in sub-tasks 1 and 3}. 

(3) \textbf{In a `Multilingual Joint' training scenario, German demonstrates a consistent preference for the adapter training technique across all three sub-tasks}. In particular, German is the only language in sub-task 2 where FFT is not producing the best performance for the `Multilingual Joint' training scenario. It is also the only case where adapter models demonstrate better performance than LoRA for seen languages for sub-task 3 in the joint training scenario.

We also observe that LoRA, in a `Multilingual Joint' training scenario, shows the best overall performance for sub-task 3. Since this method was not used by any of the teams participating in the shared task, we want to look into how this approach compares to the official leaderboard results after the competition. Table~\ref{table:st3_results} demonstrates the scores of the winning system for sub-task 3 along with the scores achieved in sub-task 3 in these experiments.
As can be seen, the LoRA method for sub-task 3 outperforms most of the results of the winning systems. We achieve an increase in the performance of up to 19.63\% for all the languages except for Georgian (KA). 6 out of 9 languages (FR, IT, PL, RU, ES and EL) achieve the best result in the `Multilingual Joint' training scenario when applying LoRA. Not surprisingly, the best score for English is achieved in one-to-one `English only' scenario. This increase also results in the first placings in 8 out of 9 languages.

\begin{table}[H]
\caption{Sub-task 3 official test set leaderboard comparison.}
\begin{tabular}{|llcccc|}
\hline
\multicolumn{1}{|l|}{Language} & \multicolumn{1}{l|}{1st Place Team} & \multicolumn{1}{l|}{\begin{tabular}[c]{@{}l@{}}\Fmicro\\ (1st Place)\end{tabular}} & \multicolumn{1}{l|}{\begin{tabular}[c]{@{}l@{}}\Fmicro\\ (Ours)\end{tabular}} & \multicolumn{1}{l|}{\begin{tabular}[c]{@{}l@{}}\Fmicro\\ Increase\end{tabular}} & \multicolumn{1}{l|}{\begin{tabular}[c]{@{}l@{}}Final Placing\\      (Ours)\end{tabular}} \\ \hline
\multicolumn{1}{|l|}{EN} & \multicolumn{1}{l|}{APatt} & \multicolumn{1}{c|}{0.37562} & \multicolumn{1}{c|}{0.44937} & \multicolumn{1}{c|}{19.63\%} & 1 \\ \hline
\multicolumn{1}{|l|}{FR} & \multicolumn{1}{l|}{NAP} & \multicolumn{1}{c|}{0.46869} & \multicolumn{1}{c|}{0.49238} & \multicolumn{1}{c|}{5.05\%} & 1 \\ \hline
\multicolumn{1}{|l|}{DE} & \multicolumn{1}{l|}{KInITVeraAI} & \multicolumn{1}{c|}{0.51304} & \multicolumn{1}{c|}{0.54174} & \multicolumn{1}{c|}{5.30\%} & 1 \\ \hline
\multicolumn{1}{|l|}{IT} & \multicolumn{1}{l|}{KInITVeraAI} & \multicolumn{1}{c|}{0.55019} & \multicolumn{1}{c|}{0.59919} & \multicolumn{1}{c|}{8.91\%} & 1 \\ \hline
\multicolumn{1}{|l|}{PL} & \multicolumn{1}{l|}{KInITVeraAI} & \multicolumn{1}{c|}{0.43037} & \multicolumn{1}{c|}{0.44964} & \multicolumn{1}{c|}{4.48\%} & 1 \\ \hline
\multicolumn{1}{|l|}{RU} & \multicolumn{1}{l|}{KInITVeraAI} & \multicolumn{1}{c|}{0.38682} & \multicolumn{1}{c|}{0.42635} & \multicolumn{1}{c|}{10.22\%} & 1 \\ \hline
\multicolumn{1}{|l|}{ES} & \multicolumn{1}{l|}{TeamAmpa} & \multicolumn{1}{c|}{0.38106} & \multicolumn{1}{c|}{0.40674} & \multicolumn{1}{c|}{6.74\%} & 1 \\ \hline
\multicolumn{1}{|l|}{EL} & \multicolumn{1}{l|}{KInITVeraAI} & \multicolumn{1}{c|}{0.26733} & \multicolumn{1}{c|}{0.27668} & \multicolumn{1}{c|}{3.38\%} & 1 \\ \hline
\multicolumn{1}{|l|}{KA} & \multicolumn{1}{l|}{KInITVeraAI} & \multicolumn{1}{c|}{0.45714} & \multicolumn{1}{c|}{0.448} & \multicolumn{1}{c|}{-2.00\%} & 2 \\ \hline
\end{tabular}
\label{table:st3_results}
\end{table}

To summarise this analysis, we observe that adapter models and LoRA produce comparable or even better results for sub-task 3 or in the scenarios when the training data is scarce or is only available in one language. This is a promising result given that these approaches require less memory and training time, as discussed in the previous section.

\section*{Limitations and discussion}
The results reported in this study are limited to using one dataset annotated with different classification tasks. Further analysis would be beneficial to test these findings on a diverse set of corpora and tasks. Additionally, we examined the effect of one adapter technique, which prevents us from generalising our findings to all types of adapter methods.
Finally, since sub-tasks 1 and 2 are performed at article-level where the average length of the article is more than 512 tokens (Table~\ref{tab:dataset-statistics}), many of the signals present in the rest of the articles could potentially be ignored by all of the methods compared in this study.

Our findings provide novel insights into the effectiveness of PEFT techniques in multilingual classification tasks. In particular,
we found that the multilingual training scenario on average improves the performance on all sub-tasks. However, the best performing methods within this setting differed depending on the sub-task. 

Our results on the LoRA method are novel as this PEFT technique was not previously investigated in multilingual tasks. While we observe that sub-tasks 1 and 2 benefit from FFT method in a `Multilingual Joint' training scenario, our results demonstrate a particularly interesting behaviour of the LoRA method in this training scenario for sub-task 3, where it consistently outperforms the FFT approach. 
While a thorough analysis is needed to conclude why this is the case, the differences in the properties of these tasks that we presented in Table 1 could potentially bring light to this question. Firstly, sub-task 3 is trained on and applied to much shorter texts than sub-tasks 1 and 2, while having substantially more training examples. Additionally, sub-task 3 is characterised by a high number of classes (23 compared to 14 in sub-task 2 and 3 in sub-task 1) and a much more severe data imbalance, as shown in Table 1. The nature of the tasks is also quite different. The task of framing detection relies less on commonsense knowledge and pragmatics, while sub-tasks 1 and 3 are more subjective for human experts. All of these factors can potentially influence the performance of each method in the respective sub-tasks, and further research is needed to make conclusions regarding the types of classification tasks that benefit from from LoRA method.

We observe the adapter method to outperform FFT only in the zero-shot cross-lingual scenario when a sufficient amount of data is provided in the source language (`English + Translations' scenario) while being much less efficient when the training data contains a variety of languages or is limited and monolingual. This is an important insight that complements previous studies  which report the adapter approach to always outperform FFT in a cross-lingual zero-shot inference \cite{he-etal-2021-effectiveness, chalkidis-etal-2021-multieurlex}. Our results also challenge a prior conclusion by Xenouleas et al. \cite{xenouleas2022realistic} which suggests that adapter methods are universally beneficial for low-resource and cross-lingual tasks. Additionally, our results disagree with the prior reports that the FFT approach is always outperformed by the adapter method in a zero-shot cross-lingual inference \cite{he-etal-2021-effectiveness, chalkidis-etal-2021-multieurlex}. Contrary to this, we found that for the `unseen' languages consistent across all training scenarios (‘ES’, ‘EL’ and ‘KA’), FFT outperforms adapter method in all training scenarios for sub-task 1 and sub-task 3, and for two our three scenarios in sub-task 2, ‘Multilingual Joint’ and ‘English Only’. Finally, our study provides the  first to our knowledge comparison of the `Multilingual Joint' scenario with the translation-based monolingual one in terms of the zero-shot cross-lingual inference.

While sub-task 3 performs better for the LoRA approach in a multilingual setting on average, this is not consistent across seen and zero-shot languages, with FFT performing better in a zero-shot cross-lingual setting. This preference is particularly strong in the cases when  the amount of training data is limited. This can indicate the fact that the adapter method is potentially not good at cross-lingual generalisation, however, more analysis is needed on a wider variety of tasks.

In the tasks with highly skewed data, adapter method can potentially be more prone than FFT to favoring the most frequent class, since as can be observed from Table~\ref{tab:main-results}, \Fmicro score for sub-task 1 is significantly higher for adapter method compared to FFT, while FFT results in a higher \Fmacro. However, a detailed error analysis to confirm or refute this assumption is currently not possible since the gold-standard labels for the test set are not released.

Low-resource languages may have a preference for FFT in all training scenarios when predictions are made in a cross-lingual zero-shot way. This assumption is suggested by the fact that Georgian is the only language that consistently prefers the FFT approach across all training scenarios and for all sub-tasks.

Interestingly, the performance on Georgian in sub-tasks 1 and 2 is higher than that on seen languages, despite being low-resource and zero-shot. One potential reason for this observation could be the fact that, as was previously shown in Tables~\ref{tab:data_st1} and \ref{tab:data_st2}, the texts in the test set for Georgian are, on average, within the limit of XLM-R and are much shorter than the input lengths for other 8 languages. This could explain why the same effect is not observed for sub-task 3, where all the inputs are within the transformer token limit and are relatively short. However, with the lack of the gold standard labels for surprise languages, it is not possible to eliminate other reasons for this phenomenon, as it could also be explained with the lack of particularly difficult to predict classes in the test set for Georgian for both sub-tasks.

\section*{Conclusion}

In this work, we performed the first (to our knowledge) analysis of the performance of Low-Rank Adaptation (LoRA) technique and its comparison with the adapter and full fine-tuning (FFT) methods in a multilingual multiclass scenario. 

We found that parameter-efficient fine-tuning techniques (PEFTs), LoRA and bottleneck adapter, provide significant computation efficiency compared to FFT in terms of the training time, the number of trainable parameters and the amount of VRAM memory required. In particular, they reduce the number of trainable parameters between 140 and 280 times and achieve between 32\% and 44\% shorter training time. 

The comparison between LoRA and adapter methods in terms of the parameter efficiency suggests that their performance depends on a certain sub-task and hyperparameters used. This observation is aligned with the results of the previous study by He et al. \cite{he-etal-2021-effectiveness}, who found the benefit of the adapter approach to be task-dependent. While we observe LoRA to be more efficient than the adapter method for tasks of news articles' genre and framing detection, the adapter method takes less average time and uses less training parameters for the latter one.

 Moreover, we found the performance of the methods to be highly dependent on the training scenario. Adapter method performs better than LoRA and FFT in the scenario where there is a lack of language diversity in the training set across the sub-tasks. 

 The differences between all three methods become insignificant, often less than 1\% on average, as the size of the training data decreases. This indicates that it is possible to achieve high computational efficiency by using PEFT methods without losing much in terms of the classification performance in this setting. More experiments on this result involving a gradual decrease in the size of the training set would be beneficial in future to find the threshold when the performances across the methods match or when PEFT methods become more classification-efficient.

The performance on the unseen languages is often highly dependent on the training scenario. We found that FFT performs better than PEFT methods in zero-shot cross-lingual predictions when trained on a joint multilingual dataset, which is different from the results reported by Chalkidis et al. \cite{chalkidis-etal-2021-multieurlex}. However, we observe the effect reported by the authors in a monolingual training scenario, where adapter method performs better on zero-shot languages.

Finally, the multilingual joint LoRA setting allowed us to significantly improve our official results on the SemEval 2023 sub-task 3 (persuasion techniques detection) and to outperform most of the official leaderboard-best results, placing first in all languages except Georgian, where we are in the second place compared to the official leaderboard results.


\section*{Supporting information}

\paragraph*{S1 Appendix.}
\label{S1_Appendix_ST3cats}
{\bf Complete list of categories and their hierarchy for sub-task 3.}
\paragraph*{S2 Appendix.}
\label{S2_Appendix_bitfit}
{\bf Ablation study for PEFT techniques.}
\paragraph*{S3 Appendix.}
\label{S3_Appendix_ROBERTA}
{\bf Model size selection for XLM-RoBERTa.}
\paragraph*{S4 Appendix}
\label{S4_Appendix_Resourse}
{\bf Amount of resources per-language.}
\paragraph*{S5 Appendix}
\label{S5_Appendix_adapter}
{\bf Comparison of Houlsby and Pfeiffer adapters across three sub-tasks.}
\section*{Acknowledgements}

We thank Joanna Wright for providing her comments and suggestions on the draft of this work.

\nolinenumbers

%
%
%






\begin{thebibliography}{10}

\bibitem{devlin2018bert}
Devlin J, Chang MW, Lee K, Toutanova K.
\newblock BERT: Pre-training of Deep Bidirectional Transformers for Language
  Understanding.
\newblock In: North American Chapter of the Association for Computational
  Linguistics; 2019.Available from:
  \url{https://api.semanticscholar.org/CorpusID:52967399}.

\bibitem{raffel2020exploring}
Raffel C, Shazeer N, Roberts A, Lee K, Narang S, Matena M, et~al.
\newblock Exploring the limits of transfer learning with a unified text-to-text
  transformer.
\newblock The Journal of Machine Learning Research. 2020;21(1):5485--5551.

\bibitem{zaken2021bitfit}
Ben~Zaken E, Goldberg Y, Ravfogel S.
\newblock {B}it{F}it: Simple Parameter-efficient Fine-tuning for
  Transformer-based Masked Language-models.
\newblock In: Muresan S, Nakov P, Villavicencio A, editors. Proceedings of the
  60th Annual Meeting of the Association for Computational Linguistics (Volume
  2: Short Papers). Dublin, Ireland: Association for Computational Linguistics;
  2022. p. 1--9.
\newblock Available from: \url{https://aclanthology.org/2022.acl-short.1}.

\bibitem{jiang2019smart}
Jiang H, He P, Chen W, Liu X, Gao J, Zhao T.
\newblock {SMART}: Robust and Efficient Fine-Tuning for Pre-trained Natural
  Language Models through Principled Regularized Optimization.
\newblock In: Jurafsky D, Chai J, Schluter N, Tetreault J, editors. Proceedings
  of the 58th Annual Meeting of the Association for Computational Linguistics.
  Online: Association for Computational Linguistics; 2020. p. 2177--2190.
\newblock Available from: \url{https://aclanthology.org/2020.acl-main.197}.

\bibitem{xu2021raise}
Xu R, Luo F, Zhang Z, Tan C, Chang B, Huang S, et~al.
\newblock Raise a Child in Large Language Model: Towards Effective and
  Generalizable Fine-tuning.
\newblock In: Moens MF, Huang X, Specia L, Yih SWt, editors. Proceedings of the
  2021 Conference on Empirical Methods in Natural Language Processing. Online
  and Punta Cana, Dominican Republic: Association for Computational
  Linguistics; 2021. p. 9514--9528.
\newblock Available from: \url{https://aclanthology.org/2021.emnlp-main.749}.

\bibitem{lialin2023scaling}
Lialin V, Deshpande V, Rumshisky A.
\newblock Scaling down to scale up: A guide to parameter-efficient fine-tuning.
\newblock arXiv preprint arXiv:230315647. 2023;.

\bibitem{li2021prefix}
Li XL, Liang P.
\newblock Prefix-Tuning: Optimizing Continuous Prompts for Generation.
\newblock In: Zong C, Xia F, Li W, Navigli R, editors. Proceedings of the 59th
  Annual Meeting of the Association for Computational Linguistics and the 11th
  International Joint Conference on Natural Language Processing (Volume 1: Long
  Papers). Online: Association for Computational Linguistics; 2021. p.
  4582--4597.
\newblock Available from: \url{https://aclanthology.org/2021.acl-long.353}.

\bibitem{lester2021power}
Lester B, Al-Rfou R, Constant N.
\newblock The Power of Scale for Parameter-Efficient Prompt Tuning.
\newblock In: Moens MF, Huang X, Specia L, Yih SWt, editors. Proceedings of the
  2021 Conference on Empirical Methods in Natural Language Processing. Online
  and Punta Cana, Dominican Republic: Association for Computational
  Linguistics; 2021. p. 3045--3059.
\newblock Available from: \url{https://aclanthology.org/2021.emnlp-main.243}.

\bibitem{bapna-firat-2019-simple}
Bapna A, Firat O.
\newblock Simple, Scalable Adaptation for Neural Machine Translation.
\newblock In: Inui K, Jiang J, Ng V, Wan X, editors. Proceedings of the 2019
  Conference on Empirical Methods in Natural Language Processing and the 9th
  International Joint Conference on Natural Language Processing (EMNLP-IJCNLP).
  Hong Kong, China: Association for Computational Linguistics; 2019. p.
  1538--1548.
\newblock Available from: \url{https://aclanthology.org/D19-1165}.

\bibitem{houlsby2019parameter}
Houlsby N, Giurgiu A, Jastrzebski S, Morrone B, De~Laroussilhe Q, Gesmundo A,
  et~al.
\newblock Parameter-Efficient Transfer Learning for {NLP}.
\newblock In: Chaudhuri K, Salakhutdinov R, editors. Proceedings of the 36th
  International Conference on Machine Learning. vol.~97 of Proceedings of
  Machine Learning Research. PMLR; 2019. p. 2790--2799.
\newblock Available from:
  \url{https://proceedings.mlr.press/v97/houlsby19a.html}.

\bibitem{he-etal-2021-effectiveness}
He R, Liu L, Ye H, Tan Q, Ding B, Cheng L, et~al.
\newblock On the Effectiveness of Adapter-based Tuning for Pretrained Language
  Model Adaptation.
\newblock In: Proceedings of the 59th Annual Meeting of the Association for
  Computational Linguistics and the 11th International Joint Conference on
  Natural Language Processing (Volume 1: Long Papers). Online: Association for
  Computational Linguistics; 2021. p. 2208--2222.
\newblock Available from: \url{https://aclanthology.org/2021.acl-long.172}.

\bibitem{hu2021lora}
Hu EJ, Shen Y, Wallis P, Allen-Zhu Z, Li Y, Wang S, et~al.
\newblock Lo{RA}: Low-Rank Adaptation of Large Language Models.
\newblock In: International Conference on Learning Representations;
  2022.Available from: \url{https://openreview.net/forum?id=nZeVKeeFYf9}.

\bibitem{chalkidis-etal-2021-multieurlex}
Chalkidis I, Fergadiotis M, Androutsopoulos I.
\newblock {M}ulti{EURLEX} - A multi-lingual and multi-label legal document
  classification dataset for zero-shot cross-lingual transfer.
\newblock In: Proceedings of the 2021 Conference on Empirical Methods in
  Natural Language Processing. Online and Punta Cana, Dominican Republic:
  Association for Computational Linguistics; 2021. p. 6974--6996.
\newblock Available from: \url{https://aclanthology.org/2021.emnlp-main.559}.

\bibitem{xenouleas2022realistic}
Xenouleas S, Tsoukara A, Panagiotakis G, Chalkidis I, Androutsopoulos I.
\newblock Realistic Zero-Shot Cross-Lingual Transfer in Legal Topic
  Classification.
\newblock In: Proceedings of the 12th Hellenic Conference on Artificial
  Intelligence. SETN '22. New York, NY, USA: Association for Computing
  Machinery; 2022.Available from:
  \url{https://doi.org/10.1145/3549737.3549760}.

\bibitem{semeval2023task3}
Piskorski J, Stefanovitch N, Da~San~Martino G, Nakov P.
\newblock {S}em{E}val-2023 Task 3: Detecting the Category, the Framing, and the
  Persuasion Techniques in Online News in a Multi-lingual Setup.
\newblock In: Proceedings of the The 17th International Workshop on Semantic
  Evaluation (SemEval-2023). Toronto, Canada: Association for Computational
  Linguistics; 2023. p. 2343--2361.
\newblock Available from: \url{https://aclanthology.org/2023.semeval-1.317}.

\bibitem{wu2023sheffieldveraai}
Wu B, Razuvayevskaya O, Heppell F, Leite JA, Scarton C, Bontcheva K, et~al.
\newblock {S}heffield{V}era{AI} at {S}em{E}val-2023 Task 3: Mono and
  Multilingual Approaches for News Genre, Topic and Persuasion Technique
  Classification.
\newblock In: Proceedings of the The 17th International Workshop on Semantic
  Evaluation (SemEval-2023). Toronto, Canada: Association for Computational
  Linguistics; 2023. p. 1995--2008.
\newblock Available from: \url{https://aclanthology.org/2023.semeval-1.275}.

\bibitem{hromadka2023kinitveraai}
Hromadka T, Smolen T, Remis T, Pecher B, Srba I.
\newblock {KI}n{ITV}era{AI} at {S}em{E}val-2023 Task 3: Simple yet Powerful
  Multilingual Fine-Tuning for Persuasion Techniques Detection.
\newblock In: Proceedings of the The 17th International Workshop on Semantic
  Evaluation (SemEval-2023). Toronto, Canada: Association for Computational
  Linguistics; 2023. p. 629--637.
\newblock Available from: \url{https://aclanthology.org/2023.semeval-1.86}.

\bibitem{gururangan2020don}
Gururangan S, Marasovi{\'c} A, Swayamdipta S, Lo K, Beltagy I, Downey D, et~al.
\newblock Don{'}t Stop Pretraining: Adapt Language Models to Domains and Tasks.
\newblock In: Jurafsky D, Chai J, Schluter N, Tetreault J, editors. Proceedings
  of the 58th Annual Meeting of the Association for Computational Linguistics.
  Online: Association for Computational Linguistics; 2020. p. 8342--8360.
\newblock Available from: \url{https://aclanthology.org/2020.acl-main.740}.

\bibitem{pfeiffer-etal-2020-mad}
Pfeiffer J, Vuli{\'c} I, Gurevych I, Ruder S.
\newblock {MAD-X}: {A}n {A}dapter-{B}ased {F}ramework for {M}ulti-{T}ask
  {C}ross-{L}ingual {T}ransfer.
\newblock In: Proceedings of the 2020 Conference on Empirical Methods in
  Natural Language Processing (EMNLP). Online: Association for Computational
  Linguistics; 2020. p. 7654--7673.
\newblock Available from: \url{https://aclanthology.org/2020.emnlp-main.617}.

\bibitem{dettmers2023qlora}
Dettmers T, Pagnoni A, Holtzman A, Zettlemoyer L.
\newblock QLoRA: Efficient finetuning of quantized llms.
\newblock arXiv preprint arXiv:230514314. 2023;.

\bibitem{wang-etal-2018-glue}
Wang A, Singh A, Michael J, Hill F, Levy O, Bowman S.
\newblock {GLUE}: A Multi-Task Benchmark and Analysis Platform for Natural
  Language Understanding.
\newblock In: Proceedings of the 2018 {EMNLP} Workshop {B}lackbox{NLP}:
  Analyzing and Interpreting Neural Networks for {NLP}. Brussels, Belgium:
  Association for Computational Linguistics; 2018. p. 353--355.
\newblock Available from: \url{https://aclanthology.org/W18-5446}.

\bibitem{yu2022beyond}
Yu X, Chatterjee T, Asai A, Hu J, Choi E.
\newblock Beyond Counting Datasets: A Survey of Multilingual Dataset
  Construction and Necessary Resources.
\newblock In: Goldberg Y, Kozareva Z, Zhang Y, editors. Findings of the
  Association for Computational Linguistics: EMNLP 2022. Abu Dhabi, United Arab
  Emirates: Association for Computational Linguistics; 2022. p. 3725--3743.
\newblock Available from:
  \url{https://aclanthology.org/2022.findings-emnlp.273}.

\bibitem{kiesel-etal-2019-semeval}
Kiesel J, Mestre M, Shukla R, Vincent E, Adineh P, Corney D, et~al.
\newblock {S}em{E}val-2019 Task 4: Hyperpartisan News Detection.
\newblock In: Proceedings of the 13th International Workshop on Semantic
  Evaluation. Minneapolis, Minnesota, USA: Association for Computational
  Linguistics; 2019. p. 829--839.
\newblock Available from: \url{https://aclanthology.org/S19-2145}.

\bibitem{abu-farha-etal-2022-semeval}
Abu~Farha I, Oprea SV, Wilson S, Magdy W.
\newblock {S}em{E}val-2022 Task 6: i{S}arcasm{E}val, Intended Sarcasm Detection
  in {E}nglish and {A}rabic.
\newblock In: Proceedings of the 16th International Workshop on Semantic
  Evaluation (SemEval-2022). Seattle, United States: Association for
  Computational Linguistics; 2022. p. 802--814.
\newblock Available from: \url{https://aclanthology.org/2022.semeval-1.111}.

\bibitem{da-san-martino-etal-2020-semeval}
Da~San~Martino G, Barr{\'o}n-Cede{\~n}o A, Wachsmuth H, Petrov R, Nakov P.
\newblock {S}em{E}val-2020 Task 11: Detection of Propaganda Techniques in News
  Articles.
\newblock In: Proceedings of the Fourteenth Workshop on Semantic Evaluation.
  Barcelona (online): International Committee for Computational Linguistics;
  2020. p. 1377--1414.
\newblock Available from: \url{https://aclanthology.org/2020.semeval-1.186}.

\bibitem{dimitrov-etal-2021-semeval}
Dimitrov D, Bin~Ali B, Shaar S, Alam F, Silvestri F, Firooz H, et~al.
\newblock {S}em{E}val-2021 Task 6: Detection of Persuasion Techniques in Texts
  and Images.
\newblock In: Proceedings of the 15th International Workshop on Semantic
  Evaluation (SemEval-2021). Online: Association for Computational Linguistics;
  2021. p. 70--98.
\newblock Available from: \url{https://aclanthology.org/2021.semeval-1.7}.

\bibitem{card-etal-2015-media}
Card D, Boydstun AE, Gross JH, Resnik P, Smith NA.
\newblock The Media Frames Corpus: Annotations of Frames Across Issues.
\newblock In: Proceedings of the 53rd Annual Meeting of the Association for
  Computational Linguistics and the 7th International Joint Conference on
  Natural Language Processing (Volume 2: Short Papers). Beijing, China:
  Association for Computational Linguistics; 2015. p. 438--444.
\newblock Available from: \url{https://aclanthology.org/P15-2072}.

\bibitem{da-san-martino-etal-2019-fine}
Da~San~Martino G, Yu S, Barr{\'o}n-Cede{\~n}o A, Petrov R, Nakov P.
\newblock Fine-Grained Analysis of Propaganda in News Article.
\newblock In: Proceedings of the 2019 Conference on Empirical Methods in
  Natural Language Processing and the 9th International Joint Conference on
  Natural Language Processing (EMNLP-IJCNLP). Hong Kong, China: Association for
  Computational Linguistics; 2019. p. 5636--5646.
\newblock Available from: \url{https://aclanthology.org/D19-1565}.

\bibitem{HHUSemeval2023task3}
Billert F, Conrad S.
\newblock {HHU} at {S}em{E}val-2023 Task 3: An Adapter-based Approach for News
  Genre Classification.
\newblock In: Proceedings of the The 17th International Workshop on Semantic
  Evaluation (SemEval-2023). Toronto, Canada: Association for Computational
  Linguistics; 2023. p. 1166--1171.
\newblock Available from: \url{https://aclanthology.org/2023.semeval-1.162}.

\bibitem{NAPSemeval2023task3}
Falk N, Eichel A, Piccirilli P.
\newblock {NAP} at {S}em{E}val-2023 Task 3: Is Less Really More?
  (Back-)Translation as Data Augmentation Strategies for Detecting Persuasion
  Techniques.
\newblock In: Proceedings of the The 17th International Workshop on Semantic
  Evaluation (SemEval-2023). Toronto, Canada: Association for Computational
  Linguistics; 2023. p. 1433--1446.
\newblock Available from: \url{https://aclanthology.org/2023.semeval-1.198}.

\bibitem{liu2019roberta}
Liu Y, Ott M, Goyal N, Du J, Joshi M, Chen D, et~al.
\newblock {RoBERTa}: A Robustly Optimized BERT Pretraining Approach.
\newblock Computing Research Repository. 2019;arXiv:1907.11692.
\newblock doi:{10.48550/ARXIV.1907.11692}.

\bibitem{annot-propaganda}
Piskorski J, Stefanovitch N, Bausier VA, Faggiani N, Linge J, Kharazi S, et~al.
\newblock News Categorization, Framing and Persuasion Techniques: Annotation
  Guidelines.
\newblock European Commission, Ispra, JRC132862; 2023.

\bibitem{barbaresi-2021-trafilatura}
Barbaresi A.
\newblock {Trafilatura: A Web Scraping Library and Command-Line Tool for Text
  Discovery and Extraction}.
\newblock In: Proceedings of the Joint Conference of the 59th Annual Meeting of
  the Association for Computational Linguistics and the 11th International
  Joint Conference on Natural Language Processing: System Demonstrations.
  Association for Computational Linguistics; 2021. p. 122--131.
\newblock Available from: \url{https://aclanthology.org/2021.acl-demo.15}.

\bibitem{PIKULIAK2021113765}
Pikuliak M, Šimko M, Bieliková M.
\newblock Cross-lingual learning for text processing: A survey.
\newblock Expert Systems with Applications. 2021;165:113765.
\newblock doi:{https://doi.org/10.1016/j.eswa.2020.113765}.

\bibitem{conneau-etal-2020-unsupervised}
Conneau A, Khandelwal K, Goyal N, Chaudhary V, Wenzek G, Guzm{\'a}n F, et~al.
\newblock Unsupervised Cross-lingual Representation Learning at Scale.
\newblock In: Proceedings of the 58th Annual Meeting of the Association for
  Computational Linguistics. Online: Association for Computational Linguistics;
  2020. p. 8440--8451.
\newblock Available from: \url{https://aclanthology.org/2020.acl-main.747}.

\bibitem{pfeiffer-etal-2020-adapterhub}
Pfeiffer J, R{\"u}ckl{\'e} A, Poth C, Kamath A, Vuli{\'c} I, Ruder S, et~al.
\newblock {A}dapter{H}ub: A Framework for Adapting Transformers.
\newblock In: Proceedings of the 2020 Conference on Empirical Methods in
  Natural Language Processing: System Demonstrations. Online: Association for
  Computational Linguistics; 2020. p. 46--54.
\newblock Available from: \url{https://aclanthology.org/2020.emnlp-demos.7}.

\bibitem{pires-etal-2019-multilingual}
Pires T, Schlinger E, Garrette D.
\newblock How Multilingual is Multilingual {BERT}?
\newblock In: Proceedings of the 57th Annual Meeting of the Association for
  Computational Linguistics. Florence, Italy: Association for Computational
  Linguistics; 2019. p. 4996--5001.
\newblock Available from: \url{https://aclanthology.org/P19-1493}.

\bibitem{savenkov-lopez-2022-state}
Savenkov K, Lopez M.
\newblock The State of the Machine Translation 2022.
\newblock In: Proceedings of the 15th Biennial Conference of the Association
  for Machine Translation in the Americas (Volume 2: Users and Providers Track
  and Government Track). Orlando, USA: Association for Machine Translation in
  the Americas; 2022. p. 32--49.
\newblock Available from: \url{https://aclanthology.org/2022.amta-upg.4}.

\bibitem{wang2020negative}
Wang Z, Lipton ZC, Tsvetkov Y.
\newblock On Negative Interference in Multilingual Models: Findings and A
  Meta-Learning Treatment.
\newblock In: Webber B, Cohn T, He Y, Liu Y, editors. Proceedings of the 2020
  Conference on Empirical Methods in Natural Language Processing (EMNLP).
  Online: Association for Computational Linguistics; 2020. p. 4438--4450.
\newblock Available from: \url{https://aclanthology.org/2020.emnlp-main.359}.

\end{thebibliography}

\label{LastPageBeforeSuppl}
\newpage
\pagenumbering{alph}
\rfoot{\thepage/\pageref{LastPage}}
\appendix
\begin{NoHyper}
\section*{S1 Appendix - Sub-task 3 Category Taxonomy} \label{s:taxonomy}

Descriptions of each category are provided in the original task paper \cite{semeval2023task3}.

\begin{itemize}
    \item \textbf{Attack on Reputation}
        \begin{itemize}
            \item Name Calling or Labelling
            \item Guilt by Association
            \item Casting Doubt
            \item Appeal to Hypocrisy
            \item Questioning the Reputation
        \end{itemize}
    \item \textbf{Justification}
        \begin{itemize}
            \item Flag Waving
            \item Appeal to Authority
            \item Appeal to Popularity
            \item Appeal to Values
            \item Appeal to Fear, Prejudice
        \end{itemize}
    \item \textbf{Distraction}
        \begin{itemize}
            \item Strawman
            \item Red Herring
            \item Whataboutism
        \end{itemize}
    \item \textbf{Simplification}
        \begin{itemize}
            \item Causal Oversimplification
            \item False Dilemma or No Choice
            \item Consequential Oversimplification
        \end{itemize}
    \item \textbf{Call}
        \begin{itemize}
            \item Slogans
            \item Conversation Killer
            \item Appeal to Time
        \end{itemize}
    \item \textbf{Manipulative Wording}
        \begin{itemize}
            \item Loaded Language
            \item Obfuscation, Intentional Vagueness, Confusion
            \item Exaggeration or Minimisation
            \item Repetition
        \end{itemize}
\end{itemize}
\end{NoHyper}
\clearpage

\section*{S2 Appendix - ablation study for PEFT techniques}\label{s:bitfit}
In this section, we perform the ablation study to find how each of three PEFT techniques performs in a multilingual, cross-lingual and monolingual scenarios on sub-task 1. As can be seen, while LoRA and Adapter often demonstrate comparable behaviour in all training scenarios, the BitFit method achieves significantly lower results. Importantly, while the results in the multilingual setting are comparable, the difference between BitFit and two other PEFTs is particularly noticeable for the two cross-lingual settings (‘English + Translations' and ‘English Only') scenarios.

\begin{table}[H]
\small
{\begin{adjustwidth}{-2.25in}{0in}
\renewcommand\thetable{S2}
\caption{\bf Sub-task 1: Genre Detection - Mean ± 1 STD \Fmacro scores for three PEFT techniques in 3 training scenarios.}
\def\arraystretch{1.1}
\begin{tabular}{|l|ccc|ccc|ccc|}
\hline
\multirow{2}{*}{Language} & \multicolumn{3}{c|}{Multilingual Joint} & \multicolumn{3}{c|}{English + Translations} & \multicolumn{3}{c|}{English Only} \\ \cline{2-10}
 & \multicolumn{1}{c|}{BitFit} & \multicolumn{1}{c|}{LoRA} & Adapter & \multicolumn{1}{c|}{BitFit} & \multicolumn{1}{c|}{LoRA} & Adapter & \multicolumn{1}{c|}{BitFit} & \multicolumn{1}{c|}{LoRA} & Adapter \\ \hline

\noalign{\vskip 4pt} \hline 
 
EN & \multicolumn{1}{c|}{{51.6±0.8}} & \multicolumn{1}{c|}{49.4±0.4} & \multicolumn{1}{c|}{\textbf{52.8±0.2}} & \multicolumn{1}{c|}{50.2±2.1} & \multicolumn{1}{c|}{52.4±0.6} & \textbf{*53.5±0.9} & \multicolumn{1}{c|}{40.3±1.7} & \multicolumn{1}{c|}{39.1±1.2} & \textbf{42.2±0.6} \\ \hline
FR & \multicolumn{1}{c|}{65.5±0.9} & \multicolumn{1}{c|}{67.4±2.3} & \textbf{67.5±0.9} & \multicolumn{1}{c|}{68.0±1.2} & \multicolumn{1}{c|}{\textbf{*69.2±1.5}} & {68.4±0.7} & \multicolumn{1}{c|}{65.4±1.5} & \multicolumn{1}{c|}{66.3±0.5} & \textbf{66.7±3.7} \\ \hline
DE & \multicolumn{1}{c|}{64.2±0.7} & \multicolumn{1}{c|}{64.8±1.2} & \textbf{*67.2±0.8} & \multicolumn{1}{c|}{64.1±1.5} & \multicolumn{1}{c|}{63.9±3.0} &  \textbf{65.8±1.2}& \multicolumn{1}{c|}{60.3±5.2} & \multicolumn{1}{c|}{\textbf{64.2±2.8}} &  63.6±0.7\\ \hline
IT & \multicolumn{1}{c|}{51.0±1.6} & \multicolumn{1}{c|}{\textbf{*53.4±1.8}} &  52.0±3.1& \multicolumn{1}{c|}{49.5±1.3} & \multicolumn{1}{c|}{52.0±1.5} & \textbf{52.9±0.8} & \multicolumn{1}{c|}{42.1±2.4} & \multicolumn{1}{c|}{\textbf{47.3±1.8}} & 44.2±1.1 \\ \hline
PL & \multicolumn{1}{c|}{63.6±1.7} & \multicolumn{1}{c|}{\textbf{*66.8±0.4}} & 65.2±1.5 & \multicolumn{1}{c|}{58.1±0.5} & \multicolumn{1}{c|}{\textbf{64.0±0.7}} & 61.8±2.4 & \multicolumn{1}{c|}{58.2±0.7} & \multicolumn{1}{c|}{\textbf{60.6±1.4}} & 59.6±3.0 \\ \hline
RU & \multicolumn{1}{c|}{53.1±0.9} & \multicolumn{1}{c|}{\textbf{*55.7±1.7}} & 52.8±0.9 & \multicolumn{1}{c|}{51.4±1.2} & \multicolumn{1}{c|}{54.2±0.8} & \textbf{54.9±2.5} & \multicolumn{1}{c|}{43.7±1.8} & \multicolumn{1}{c|}{\textbf{49.4±2.4}} & 48.7±0.6 \\ \Cline{1pt}{1-10}

Average  & \multicolumn{1}{c|}{58.13±1.6} & \multicolumn{1}{c|}{\textbf{*59.6±1.9}} & \textbf{*59.6±3.1} & \multicolumn{1}{c|}{56.9±2.6} & \multicolumn{1}{c|}{59.3±5.2} & \textbf{*59.6±2.7} & \multicolumn{1}{c|}{51.3±3.8} & \multicolumn{1}{c|}{54.5±6.1} & \textbf{54.2±2.9}\\ \hline
\noalign{\vskip 4pt} \hline 

ES & \multicolumn{1}{c|}{42.4±1.6} & \multicolumn{1}{c|}{41.8±0.5} & \textbf{44.2±0.7} & \multicolumn{1}{c|}{44.0±1.8} & \multicolumn{1}{c|}{46.0±1.3} & \textbf{*46.2±5} & \multicolumn{1}{c|}{40.3±0.7} & \multicolumn{1}{c|}{\textbf{41.7±0.8}} & 40.9±1.1\\ \hline
EL & \multicolumn{1}{c|}{40.1±1.3} & \multicolumn{1}{c|}{\textbf{41.4±2.7}} & 40.9±1.7 & \multicolumn{1}{c|}{39.1±1.1} & \multicolumn{1}{c|}{\textbf{*42.9±1.7}} & 42.2±3.6 & \multicolumn{1}{c|}{\textbf{36.7±1.3}} & \multicolumn{1}{c|}{38.6±1.7} & 37.5±0.8 \\ \hline
KA & \multicolumn{1}{c|}{78.5±2.2} & \multicolumn{1}{c|}{\textbf{*80.8±5.0}} & 79.2±1.8 & \multicolumn{1}{c|}{77.7±2.4} & \multicolumn{1}{c|}{\textbf{79.6±4.1}} & 77.5±2.2 & \multicolumn{1}{c|}{71.3±2.9} & \multicolumn{1}{c|}{72.9±1.5} & \textbf{74.8±2.4} \\ \Cline{1pt}{1-10}

Average & \multicolumn{1}{c|}{53.6±3.1} & \multicolumn{1}{c|}{54.7±2.4} & \textbf{54.8±1.8} & \multicolumn{1}{c|}{53.2±2.4} & \multicolumn{1}{c|}{\textbf{*56.2±3.3}} & 55.3±1.9 & \multicolumn{1}{c|}{49.2±2.7} & \multicolumn{1}{c|}{\textbf{51.1±4.0}} &  \textbf{51.1±3.3}\\ \hline
\noalign{\vskip 4pt} \hline 

\noalign{\vskip 4pt} \hline 

All & \multicolumn{1}{c|}{56.3±4.1} & \multicolumn{1}{c|}{57.9±6.3} & \textbf{58.0±2.0} & \multicolumn{1}{c|}{55.2±3.7} & \multicolumn{1}{c|}{\textbf{*58.2±3.8}} & 58.1±4.1 & \multicolumn{1}{c|}{50.6±4.4} & \multicolumn{1}{c|}{\textbf{53.3±2.6}} &  53.1±3.6 \\ \hline
\end{tabular}
\begin{flushleft}
    Best scores by language are marked with an asterisk ($*$). Best scores by training method for each training scenario are in \textbf{bold}.
\end{flushleft}
\label{table_s2}
\end{adjustwidth}}
\end{table}

\clearpage
\section*{S3 Appendix - Model size selection for XLM-RoBERTa} \label{s:roberta}

The results of our comparison between different sizes of RoBERTa model are provided in S3.1, S3.2, S3.3 Tables. As can be seen, for all training techniques, the average results are consistently significantly higher for the large size of the model for all three sub-tasks. 

\begin{table}[H] 
\begin{adjustwidth}{-2.25in}{0in}
\centering
\renewcommand\thetable{S3.1}
\caption{\bf{Comparison of the Base and Large sizes of the model in the `Multilingual Joint' scenario for the FFT method}}
\begin{tabular}{|l|rr|rr|rr|}
\hline
\multicolumn{1}{|c|}{\multirow{2}{*}{Language}} & \multicolumn{2}{c|}{Sub-task 1}                                  & \multicolumn{2}{c|}{Sub-task 2}                                  & \multicolumn{2}{c|}{Sub-task 3}                                  \\ \cline{2-7} 
\multicolumn{1}{|c|}{}                          & \multicolumn{1}{l|}{XLMR-Base} & \multicolumn{1}{l|}{XLMR-Large} & \multicolumn{1}{l|}{XLMR-Base} & \multicolumn{1}{l|}{XLMR-Large} & \multicolumn{1}{l|}{XLMR-Base} & \multicolumn{1}{l|}{XLMR-Large} \\ \hline
EN                                              & \multicolumn{1}{r|}{35.2±1.8}  & 52.7±0.5                        & \multicolumn{1}{r|}{50.9±1.1}  & 55.8±0.2                        & \multicolumn{1}{r|}{26.4±0.8}  & 34.9±1.7                        \\ \hline
FR                                              & \multicolumn{1}{r|}{69.7±0.6}  & 69.7±1.2                        & \multicolumn{1}{r|}{44.4±2.5}  & 53.3±3.3                        & \multicolumn{1}{r|}{34.2±1.4}  & 45.9±1.2                        \\ \hline
DE                                              & \multicolumn{1}{r|}{67.2±2.2}  & 66.3±0.5                        & \multicolumn{1}{r|}{60.7±1.0}  & 63.1±1.9                        & \multicolumn{1}{r|}{41.4±1.3}  & 52.1±1.9                        \\ \hline
IT                                              & \multicolumn{1}{r|}{44.5±2.1}  & 52.2±1.4                        & \multicolumn{1}{r|}{53.4±2.5}  & 59.9±1.9                        & \multicolumn{1}{r|}{46.2±0.7}  & 55.1±2.5                        \\ \hline
PL                                              & \multicolumn{1}{r|}{68.7±3.0}  & 69.2±1.1                        & \multicolumn{1}{r|}{60.1±3.5}  & 65.2±0.8                        & \multicolumn{1}{r|}{27.6±1.2}  & 40.4±3.2                        \\ \hline
RU                                              & \multicolumn{1}{r|}{55.2±1.4}  & 57.4±0.6                        & \multicolumn{1}{r|}{42.2±3.0}  & 45.3±3.0                        & \multicolumn{1}{r|}{29.6±2.1}  & 40.9±1.4                        \\ \Cline{1pt}{1-7}
ES                                              & \multicolumn{1}{r|}{40.8±2.4}  & 47.1±1.4                        & \multicolumn{1}{r|}{51.9±1.7}  & 52.7±2.1                        & \multicolumn{1}{r|}{28.3±1.4}  & 37.7±2.3                        \\ \hline
EL                                              & \multicolumn{1}{r|}{43.3±3.9}  & 40.8±2.4                        & \multicolumn{1}{r|}{48.9±1.1}  & 54.9±1.7                        & \multicolumn{1}{r|}{21.1±2.7}  & 26.7±0.9                        \\ \hline
KA                                              & \multicolumn{1}{r|}{77.5±1.8}  & 83.3±2.1                        & \multicolumn{1}{r|}{49.7±2.6}  & 60.1±4.2                        & \multicolumn{1}{r|}{33.6±2.7}  & 42.6±1.0                        \\ \Cline{1pt}{1-7}
all                                             & \multicolumn{1}{r|}{55.2±6.1}  & 59.9±3.1                        & \multicolumn{1}{r|}{51.4±6.2}  & 56.7±6.1                        & \multicolumn{1}{r|}{32.1±7.8}  & 41.8±8.6                        \\ \hline
\end{tabular}
\end{adjustwidth}
\label{tab:s31}
\end{table}

\begin{table}[H] 
\begin{adjustwidth}{-2.25in}{0in}
\centering
\renewcommand\thetable{S3.2}
\caption{\bf{Comparison of the Base and Large sizes of the model in the `Multilingual Joint' scenario for LoRA method}}
\begin{tabular}{|l|rr|rr|rr|}
\hline
\multicolumn{1}{|c|}{\multirow{2}{*}{Language}} & \multicolumn{2}{c|}{Sub-task 1}                                  & \multicolumn{2}{c|}{Sub-task 2}                                  & \multicolumn{2}{c|}{Sub-task 3}                                  \\ \cline{2-7} 
\multicolumn{1}{|c|}{}                          & \multicolumn{1}{l|}{XLMR-Base} & \multicolumn{1}{l|}{XLMR-Large} & \multicolumn{1}{l|}{XLMR-Base} & \multicolumn{1}{l|}{XLMR-Large} & \multicolumn{1}{l|}{XLMR-Base} & \multicolumn{1}{l|}{XLMR-Large} \\ \hline
EN                                              & \multicolumn{1}{r|}{45.2±2.6}  &         49.4±0.4                & \multicolumn{1}{r|}{44.9±3.4}  &  52.2±1.7                       & \multicolumn{1}{r|}{31.4±0.8}  &    37.7±0.9                     \\ \hline
FR                                              & \multicolumn{1}{r|}{62.1±0.9}  &           67.4±2.3              & \multicolumn{1}{r|}{43.8±0.5}  &   47.3±1.5                      & \multicolumn{1}{r|}{42.6±1.1}  &     48.6±0.8                    \\ \hline
DE                                              & \multicolumn{1}{r|}{60.2±1.6}  &     64.8±1.2                   & \multicolumn{1}{r|}{50.8±1.4}  &  62.3±2.2                      & \multicolumn{1}{r|}{48.2±1.3}  &   52.3±0.9                       \\ \hline
IT                                              & \multicolumn{1}{r|}{49.4±1.5}  &          53.4±1.8               & \multicolumn{1}{r|}{54.1±1.2}  &   56.8±1.6                      & \multicolumn{1}{r|}{50.4±1.9}  &    58.7±0.5                     \\ \hline
PL                                              & \multicolumn{1}{r|}{60.2±2.0}  &         66.8±0.4                & \multicolumn{1}{r|}{56.7±1.8}  &   61.0±0.8                      & \multicolumn{1}{r|}{40.3±0.7}  &     42.1±0.6                    \\ \hline
RU                                              & \multicolumn{1}{r|}{52.1±1.5}  &        55.7±1.7                 & \multicolumn{1}{r|}{35.7±0.7}  &   43.6±0.6                      & \multicolumn{1}{r|}{33.4±1.4}  &    42.3±0.3                     \\ \Cline{1pt}{1-7}
ES                                              & \multicolumn{1}{r|}{33.5±1.7}  &           41.8±0.5              & \multicolumn{1}{r|}{46.5±2.1}  & 51.7±3.0                        & \multicolumn{1}{r|}{44.2±1.2}  &       39.0±1.6                  \\ \hline
EL                                              & \multicolumn{1}{r|}{37.8±0.7}  &         41.4±2.7                & \multicolumn{1}{r|}{44.9±2.1}  &   51.4±1.2                      & \multicolumn{1}{r|}{23.6±2.2}  &    25.5±0.6                     \\ \hline
KA                                              & \multicolumn{1}{r|}{75.1±3.3}  &   80.8±5.0                     & \multicolumn{1}{r|}{48.1±2.5}  &  53.9±4.8                       & \multicolumn{1}{r|}{37.9±2.4}  &    40.4±3.1                     \\ \Cline{1pt}{1-7}
all                                             & \multicolumn{1}{r|}{52.8±4.5}  &       57.9±6.3                  & \multicolumn{1}{r|}{47.5±2.8}  &   53.4±6.0                      & \multicolumn{1}{r|}{39.7±4.7}  &       42.9±9.5                   \\ \hline
\end{tabular}
\end{adjustwidth}
\label{tab:s32}
\end{table}

\begin{table}[H] 
\begin{adjustwidth}{-2.25in}{0in}
\centering
\renewcommand\thetable{S3.3}
\caption{\bf{Comparison of the Base and Large sizes of the model in the `Multilingual Joint' scenario for the adapter method}}
\begin{tabular}{|l|rr|rr|rr|}
\hline
\multicolumn{1}{|c|}{\multirow{2}{*}{Language}} & \multicolumn{2}{c|}{Sub-task 1}                                  & \multicolumn{2}{c|}{Sub-task 2}                                  & \multicolumn{2}{c|}{Sub-task 3}                                  \\ \cline{2-7} 
\multicolumn{1}{|c|}{}                          & \multicolumn{1}{l|}{XLMR-Base} & \multicolumn{1}{l|}{XLMR-Large} & \multicolumn{1}{l|}{XLMR-Base} & \multicolumn{1}{l|}{XLMR-Large} & \multicolumn{1}{l|}{XLMR-Base} & \multicolumn{1}{l|}{XLMR-Large} \\ \hline
EN                                              & \multicolumn{1}{r|}{44.5±1.8}  &       52.8±0.2               & \multicolumn{1}{r|}{53.4±2.5}  &  55.7±2.0                       & \multicolumn{1}{r|}{30.6±3.1}  &  37.5±2.9                     \\ \hline
FR                                              & \multicolumn{1}{r|}{58.8±2.4}  &      67.5±0.9                   & \multicolumn{1}{r|}{46.4±1.2}  &   50.8±3.6                     & \multicolumn{1}{r|}{37.2±4.1}  &   45.7±1.9                     \\ \hline
DE                                              & \multicolumn{1}{r|}{61.7±2.1}  &            67.2±0.8             & \multicolumn{1}{r|}{55.0±1.1}  &  64.2±1.0                    & \multicolumn{1}{r|}{38.3±1.9}  &      53.0±0.7                   \\ \hline
IT                                              & \multicolumn{1}{r|}{43.9±3.5}  &  52.0±3.1                       & \multicolumn{1}{r|}{56.6±1.7}  &  58.2±1.0                       & \multicolumn{1}{r|}{50.4±2.2}  &         58.1±1.6               \\ \hline
PL                                              & \multicolumn{1}{r|}{59.2±1.5}  &     65.2±1.5                     & \multicolumn{1}{r|}{61.1±3.0}  &  64.1±1.7                       & \multicolumn{1}{r|}{35.2±2.7}  &       41.9±2.7                 \\ \hline
RU                                              & \multicolumn{1}{r|}{42.2±1.8}  &      52.8±0.9                   & \multicolumn{1}{r|}{37.5±1.2}  & 41.7±2.0                         & \multicolumn{1}{r|}{28.9±3.1}  &     39.2±2.6                   \\ \Cline{1pt}{1-7}
ES                                              & \multicolumn{1}{r|}{39.8±2.7}  &    44.2±0.7                     & \multicolumn{1}{r|}{40.6±3.3}  & 49.1±2.0                        & \multicolumn{1}{r|}{28.3±1.5}  &     36.7±1.3                    \\ \hline
EL                                              & \multicolumn{1}{r|}{35.7±2.1}  &     40.9±1.7                    & \multicolumn{1}{r|}{44.2±1.4}  &  54.1±2.9                       & \multicolumn{1}{r|}{21.6±1.7}  &     25.4±1.5                    \\ \hline
KA                                              & \multicolumn{1}{r|}{71.5±2.2}  &           79.2±1.8             & \multicolumn{1}{r|}{47.3±2.4}  &   55.3±1.6                      & \multicolumn{1}{r|}{37.1±1.6}  &      42.2±2.4                   \\ \Cline{1pt}{1-7}
all                                             & \multicolumn{1}{r|}{51.5±3.1}  &            58.0±2.0             & \multicolumn{1}{r|}{49.4±5.2}  &  54.8±7.1                       & \multicolumn{1}{r|}{34.8±6.6}  &     42.2±9.5                    \\ \hline
\end{tabular}
\end{adjustwidth}
\label{tab:s33}
\end{table}

\clearpage
\section*{S4 Appendix - Amount of resources per-language} 
S3 Table shows the amount of resources we considered when identifying low-resource languages for our task. For the amount of pre-training data, we summarised the statistics from \cite{conneau-etal-2020-unsupervised}. We also considered the existence of the data for fine-tuning in a certain language or whether training data is available in a language from the same group (column named "Language family"). Based on our criteria, Georgian language is a clear outlier with substantially less pre-training data and no training data in that language.
\begin{table}[H]
\begin{adjustwidth}{-2.25in}{0in}
\centering
\renewcommand\thetable{S4}
\caption{\bf{Comparison of the amount of resources per each language in sub-tasks 1, 2, and 3}}
\begin{tabular}{|l|r|rrr|c|}
\hline
\multicolumn{1}{|c|}{\multirow{2}{*}{Language}} & \multicolumn{1}{c|}{\multirow{2}{*}{\begin{tabular}[c]{@{}c@{}}Pre-training data\\ (number of tokens)\end{tabular}}} & \multicolumn{3}{c|}{\begin{tabular}[c]{@{}c@{}}Training data\\ (number of examples)\end{tabular}} & \multirow{2}{*}{Language family} \\ \cline{3-5}
\multicolumn{1}{|c|}{}                          & \multicolumn{1}{c|}{}                                                                                                & \multicolumn{1}{l|}{sub-task1}  & \multicolumn{1}{l|}{sub-task2} & \multicolumn{1}{l|}{sub-task3} &                                  \\ \hline
EN                                              & 55,608                                                                                                               & \multicolumn{1}{r|}{433}        & \multicolumn{1}{r|}{433}       & 3,610                          & West-Germanic                    \\ \hline
FR                                              & 9,780                                                                                                                & \multicolumn{1}{r|}{158}        & \multicolumn{1}{r|}{158}       & 1,693                          & Romance                          \\ \hline
DE                                              & 10,297                                                                                                               & \multicolumn{1}{r|}{132}        & \multicolumn{1}{r|}{132}       & 1,251                          & West-Germanic                    \\ \hline
IT                                              & 4,983                                                                                                                & \multicolumn{1}{r|}{226}        & \multicolumn{1}{r|}{227}       & 1,742                          & Romance                          \\ \hline
PL                                              & 6,490                                                                                                                & \multicolumn{1}{r|}{144}        & \multicolumn{1}{r|}{145}       & 1,228                          & Slavic                           \\ \hline
RU                                              & 23,408                                                                                                               & \multicolumn{1}{r|}{142}        & \multicolumn{1}{r|}{143}       & 1,232                          & Slavic                           \\ \hline
ES                                              & 9,374                                                                                                                & \multicolumn{1}{r|}{0}          & \multicolumn{1}{r|}{0}         & 0                              & Romance                          \\ \hline
EL                                              & 4,285                                                                                                                & \multicolumn{1}{r|}{0}          & \multicolumn{1}{r|}{0}         & 0                              & Hellenic                         \\ \hline
KA                                              & 469                                                                                                                  & \multicolumn{1}{r|}{0}          & \multicolumn{1}{r|}{0}         & 0                              & Kartvelian                       \\ \hline
\end{tabular}
\end{adjustwidth}
\end{table}

\clearpage
\section*{S5 Appendix - Comparison of Houlsby and Pfeiffer adapters across three sub-tasks} 
S4 Table shows the comparison of Housby and Pfeiffer configuration of the adapter for each sub-task in our experiments in a 'Multilingual Joint' training scenario for the FFT method. As can be seen, the differences are not consistent across all the languages, but on average, each sub-task benefits from the Pfeiffer adapter setting.

\begin{table}[H]
\begin{adjustwidth}{-2.25in}{0in}
\centering
\renewcommand\thetable{S5}
\caption{\bf{Comparison of the Houlsby and Pfeiffer adapters for XLM-RoBERTa Large in the `Multilingual Joint' scenario}}
\begin{tabular}{|l|rr|rr|rr|}
\hline
\multicolumn{1}{|c|}{\multirow{2}{*}{Language}} & \multicolumn{2}{c|}{Sub-task 1}          & \multicolumn{2}{l|}{Sub-task 2}          & \multicolumn{2}{l|}{Sub-task 3}          \\ \cline{2-7} 
\multicolumn{1}{|c|}{}                          & \multicolumn{1}{l|}{Housby}   & Pfeiffer & \multicolumn{1}{l|}{Houlsby}  & Pfeiffer & \multicolumn{1}{l|}{Houlsby}  & Pfeiffer \\ \hline
EN                                              & \multicolumn{1}{r|}{43.2±5.2} & 52.8±0.2 & \multicolumn{1}{r|}{54.3±2.3} & 55.7±2.0 & \multicolumn{1}{r|}{36.2±2.3} & 37.5±2.9 \\ \hline
FR                                              & \multicolumn{1}{r|}{66.7±1.3} & 67.5±0.9 & \multicolumn{1}{r|}{47.4±4.8} & 50.8±3.6 & \multicolumn{1}{r|}{47.7±1.0} & 45.7±1.9 \\ \hline
DE                                              & \multicolumn{1}{r|}{68.1±0.5} & 67.2±0.8 & \multicolumn{1}{r|}{63.5±2.1} & 64.2±1.0 & \multicolumn{1}{r|}{52.6±0.4} & 53.0±0.7 \\ \hline
IT                                              & \multicolumn{1}{r|}{50.7±2.4} & 52.0±3.1 & \multicolumn{1}{r|}{59.0±1.1} & 58.2±1.0 & \multicolumn{1}{r|}{55.1±1.3} & 58.1±1.6 \\ \hline
PL                                              & \multicolumn{1}{r|}{64.0±3.1} & 65.2±1.5 & \multicolumn{1}{r|}{62.6±0.8} & 64.1±1.7 & \multicolumn{1}{r|}{42.5±0.7} & 41.9±2.7 \\ \hline
RU                                              & \multicolumn{1}{r|}{59.7±1.9} & 52.8±0.9 & \multicolumn{1}{r|}{40.0±2.2} & 41.7±2.0 & \multicolumn{1}{r|}{40.5±0.6} & 39.2±2.6 \\ \Cline{1pt}{1-7}
ES                                              & \multicolumn{1}{r|}{47.2±1.7} & 44.2±0.7 & \multicolumn{1}{r|}{50.4±2.1} & 49.1±2.0 & \multicolumn{1}{r|}{38.5±2.2} & 36.7±1.3 \\ \hline
EL                                              & \multicolumn{1}{r|}{44.5±3.6} & 40.9±1.7 & \multicolumn{1}{r|}{52.0±2.7} & 54.1±2.9 & \multicolumn{1}{r|}{27.6±0.8} & 25.4±1.5 \\ \hline
KA                                              & \multicolumn{1}{r|}{76.1±0.4} & 79.2±1.8 & \multicolumn{1}{r|}{56.5±3.1} & 55.3±1.6 & \multicolumn{1}{r|}{43.0±2.7} & 42.2±2.4 \\ \Cline{1pt}{1-7}
all                                             & \multicolumn{1}{r|}{57.9±5.5} & 58.0±2.0 & \multicolumn{1}{r|}{54.0±8.0} & 54.8±7.1 & \multicolumn{1}{r|}{42.1±8.6} & 42.2±9.5 \\ \hline
\end{tabular}
\end{adjustwidth}

\end{table}

\end{document}